%% file: main.tex
\documentclass[10pt,twocolumn,letterpaper]{article}

\usepackage[pagenumbers]{cvpr} % To force page numbers, e.g. for an arXiv version

\input{preamble}
\usepackage{makecell}
\definecolor{cvprblue}{rgb}{0.21,0.49,0.74}
\usepackage[pagebackref,breaklinks,colorlinks,allcolors=cvprblue]{hyperref}

\def\paperID{35} % *** Enter the Paper ID here 7567
\def\confName{CVPR}
\def\confYear{2025}

\title{MirrorVerse: Pushing Diffusion Models to Realistically Reflect the World}

\author{}
\author{
    Ankit Dhiman$^{1,2}$\thanks{} \quad Manan Shah$^{1}$\footnotemark[1] \quad R Venkatesh Babu$^1$  \\ \\
    $^1$Vision and AI Lab, IISc Bangalore \quad $^2$Samsung R \& D Institute India - Bangalore
}

\begin{document}
% \maketitle

\input{sec/0_teaser}
\input{sec/0_abstract}    
\input{sec/1_intro}
\input{sec/2_related_work}
\input{sec/3_method}

\input{sec/4_results}

\input{sec/5_conclusion}

\needspace{100\baselineskip}
% %\pagebreak
% {
%     \small
%     \bibliographystyle{ieeenat_fullname}
%     \bibliography{main}
% }

% % WARNING: do not forget to delete the supplementary pages from your submission 
% % \input{sec/X_suppl}

% \end{document}

{
    \small
    \bibliographystyle{ieeenat_fullname}
    \bibliography{main}
}

%Set Counter to be 1 for the supplementary of camera-ready
\newpage
\input{sec/X_suppl}

\end{document}

%% file: preamble.tex
\usepackage{adjustbox}

\newcommand{\olddataset}{SynMirror\xspace}
\newcommand{\oldtestset}{MirrorBench\xspace}
\newcommand{\datasetname}{SynMirrorV2\xspace}
\newcommand{\testsetname}{MirrorBenchV2\xspace}
\newcommand{\oldmethod}{MirrorFusion\xspace}
\newcommand{\methodname}{MirrorFusion 2.0\xspace}

\usepackage{algorithm}
\usepackage{algorithmic}
\usepackage{needspace}
\usepackage[section]{placeins}
\usepackage{pifont}

\newcommand{\cmark}{\ding{51}}%
\newcommand{\xmark}{\ding{55}}%

%% file: sec/0_teaser.tex
\twocolumn[{%
		\renewcommand\twocolumn[1][]{#1}%
		\maketitle
            \vspace{-3em}
		\begin{center}
			\includegraphics[width=0.98\textwidth]{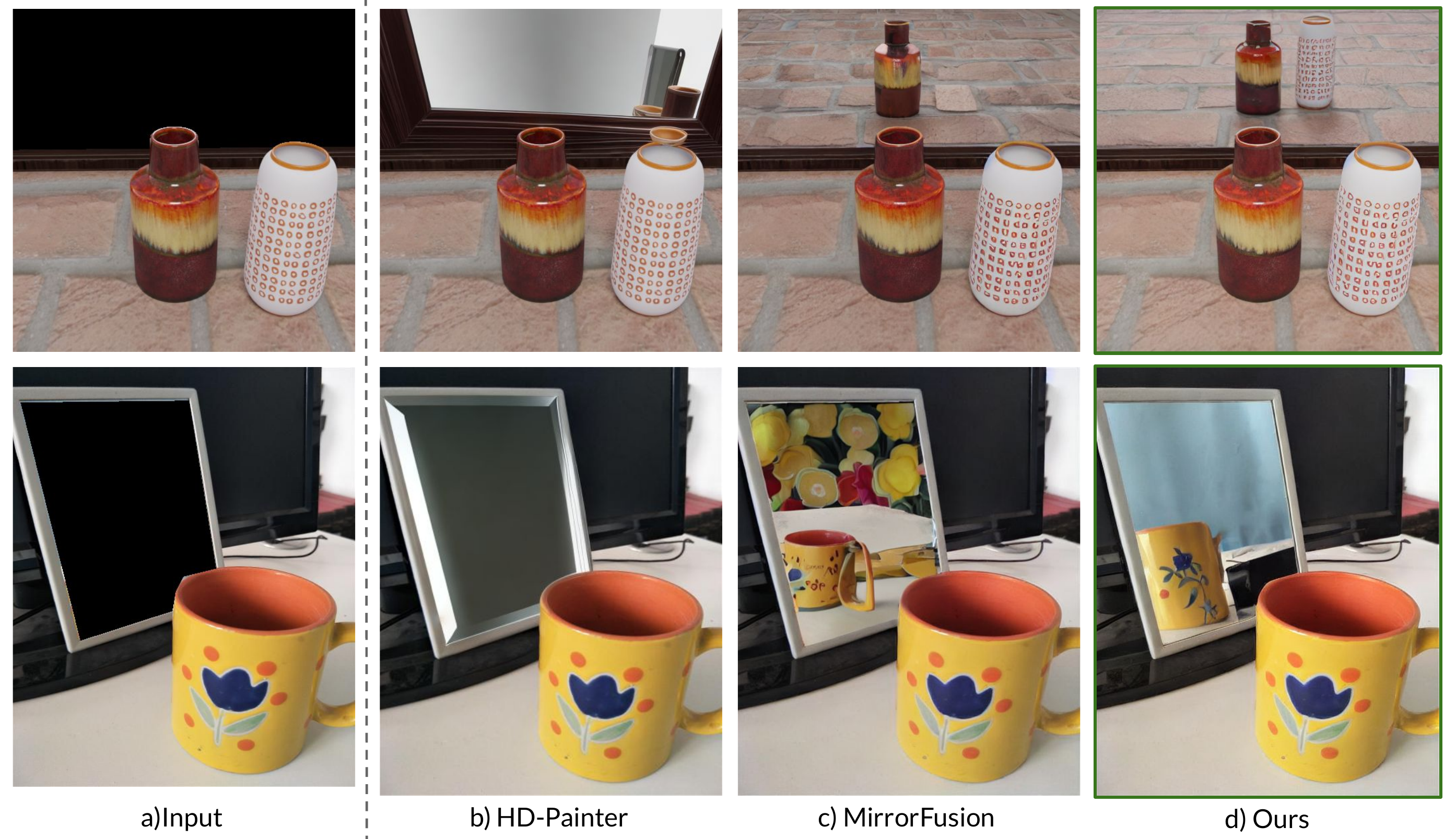}
\vspace{-0.5em}
   \captionsetup{type=figure}
			\captionof{figure}{
Our model~\methodname{}, trained on our enhanced dataset~\datasetname{} surpasses previous state-of-the-art diffusion-based inpainting models at the task of generating mirror reflections. All images were created by appending the prompt: ``A perfect plane mirror reflection of '' to the object description. All text prompts can be found in the supplementary.
}

			\label{fig:teaser}
		\end{center}
            \vspace{-1mm}
	}]
\let\thefootnote\relax\footnotetext{*Equal Contribution.}

%% file: sec/0_abstract.tex
\begin{abstract}

Diffusion models have become central to various image editing tasks, yet they often fail to fully adhere to physical laws, particularly with effects like shadows, reflections, and occlusions. In this work, we address the challenge of generating photorealistic mirror reflections using diffusion-based generative models. Despite extensive training data, existing diffusion models frequently overlook the nuanced details crucial to authentic mirror reflections. Recent approaches have attempted to resolve this by creating synthetic datasets and framing reflection generation as an inpainting task; however, they struggle to generalize across different object orientations and positions relative to the mirror. Our method overcomes these limitations by introducing key augmentations into the synthetic data pipeline: (1) random object positioning, (2) randomized rotations, and (3) grounding of objects, significantly enhancing generalization across poses and placements. To further address spatial relationships and occlusions in scenes with multiple objects, we implement a strategy to pair objects during dataset generation, resulting in a dataset robust enough to handle these complex scenarios. Achieving generalization to real-world scenes remains a challenge, so we introduce a three-stage training curriculum to develop the~\methodname{} model to improve real-world performance. We provide extensive qualitative and quantitative evaluations to support our approach.
%, and the code and data will be released for research purposes.
The project page is available at:~\href{https://mirror-verse.github.io/}{https://mirror-verse.github.io/}.

\end{abstract}

\newpage

%% file: sec/1_intro.tex
\addtocontents{toc}{\protect\setcounter{tocdepth}{-2}}
\section{Introduction}
\label{sec:intro}

In recent years, diffusion-based generative models have redefined what is possible in fields spanning from image generation to video synthesis, producing impressive results across various applications~\cite{ho2020denoising, ho2022video,podell2023sdxl, rombach2022high,esser2021taming,flux}. The evolution of these models has been accompanied by a range of methods designed to fine-tune the generation process through conditional inputs, such as edge maps, sketches, depth maps, and segmentation maps~\cite{ye2023ip,controlnet,zhao2024uni,mo2024freecontrol}. However, there remains a significant gap in their capacity to replicate intricate physical effects—particularly those rooted in the subtlety of real-world physics, including shadows~\cite{sarkar2024shadows}, specular reflections~\cite{winter2024objectdrop}, and perspective cues~\cite{upadhyay2023enhancing}. More challenging still, these techniques struggle to authentically generate mirror reflections, a task requiring a nuanced understanding of light, geometry, and realism that current methods do not adequately address. In this work, we address the question: ``\textit{Can current methods be fine-tuned to generate plausible mirror reflections?}''

\begin{figure}[!t]
    \centering
    \includegraphics[width=\linewidth]{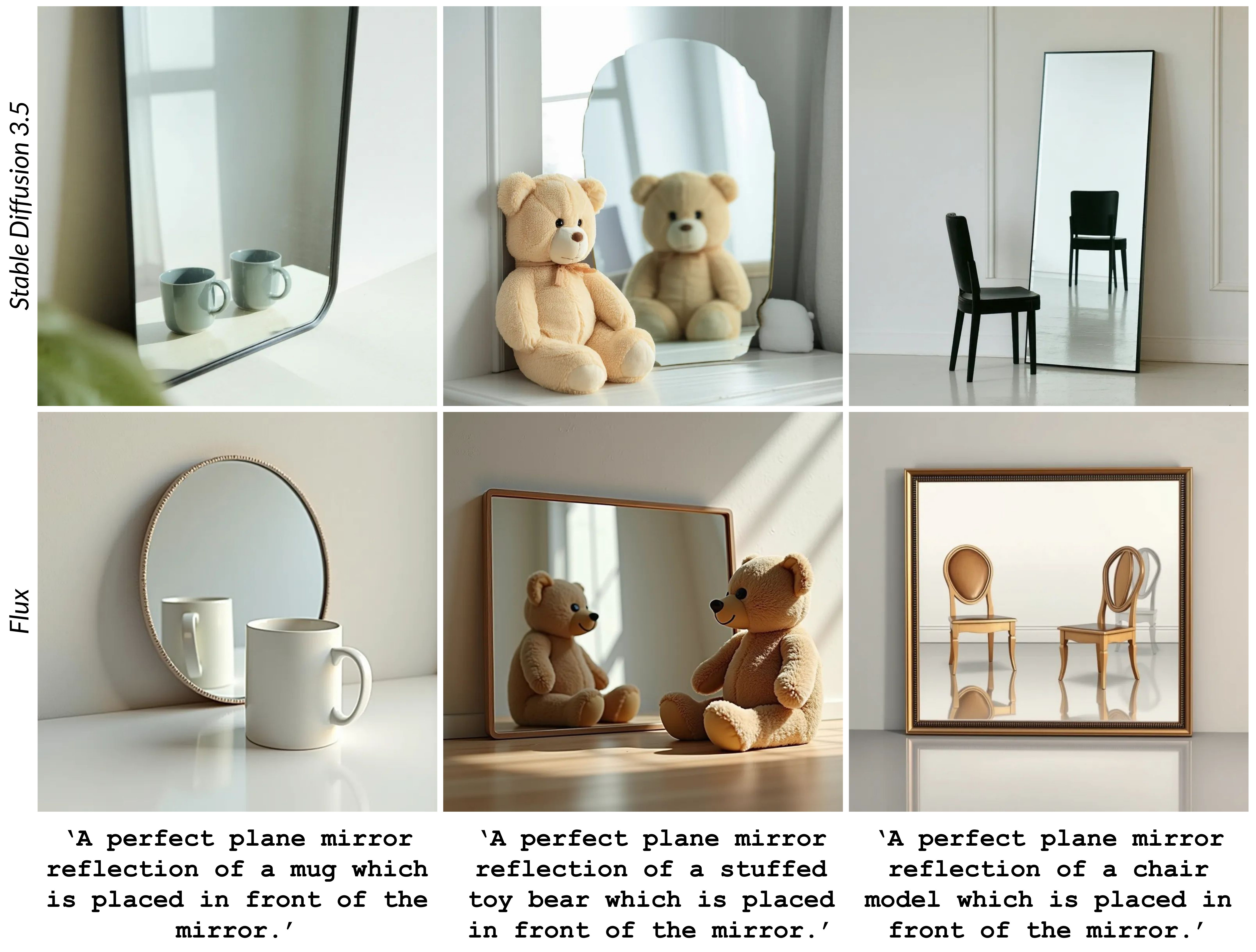}
    \caption{We observe that current state-of-the-art T2I models, SD3.5~\cite{sd35} (top row) and Flux~\cite{flux} (bottom row), face significant challenges in producing consistent and geometrically accurate reflections when prompted to generate reflections in the scene.}
    \label{fig:motivation}
    \vspace{-6mm}
\end{figure}
We motivate the problem further by providing generations from current text-to-image (T2I) generation models. We prompt Stable Diffusion 3.5~\cite{sd35} and FLUX~\cite{flux} with prompts to generate a scene with a mirror reflection.~\cref{fig:motivation} shows that these methods fail to generate plausible mirror reflections. Specifically, check the reflection of ``teddy-bear'' in the generated outputs from both the methods. Further, inpainting methods like HD-Painter~\cite{manukyan2023hd} also fail for this task, as shown in~\cref{fig:teaser}. A contemporary method called MirrorFusion~\cite{reflectingreality}, claiming to generate mirror reflections, falls short on real-world and challenging scenes as apparent in~\cref{fig:teaser}.

% Para3 Structure:
% Pain point of not having enough data with mirror and its reflectinos. Introduce our dataset and describe the augmentations. 
Despite their impressive capabilities, powerful diffusion models struggle to generate mirror reflections accurately. This limitation stems from the models' reliance on poorly learned priors, a consequence of the quality and quantity of their training data. The scarcity of high-quality, real-world images featuring mirrors and their reflections, as evidenced in ~\cref{tab:data_characteristic}, poses a significant challenge. While recent work~\cite{reflectingreality} has attempted to address this issue by training on a synthetic dataset, the results, as illustrated in ~\cref{fig:teaser}, suggest that the method's performance suffers in complex scenes and real-world settings. We hypothesize that this is due to inherent limitations in the synthetic data generation process and the training dynamics of the model.

%Para4
%Pain point : just datast is not enough. augmentations are very important.
To address the shortcomings in the synthetic data generation pipeline, we create an enhanced pipeline incorporating useful augmentations such as randomizing object position and rotation. We also ensure that the objects are anchored to the ground level in the 3D world. We observe that this diverse data improves the generalization of a trained model across the pose and position of objects in the scene. However, it does not generalize to more complex scenes with multiple objects. To address this, we propose a novel pipeline that places multiple objects in the scene based on their semantic categories, further enhancing the quality and utility of the proposed synthetic dataset. Drawing inspiration from previous works, such as those that have improved the generation quality on various tasks, notably image-editing~\cite{michel2024object}, multilingual T2I generation~\cite{li2024hunyuandita,ye2023altdiffusionamultilingual,wu2024taiyidiffusionxl} and several others, we aim to leverage the stage-wise training approach that enhanced the results in these methods.

We briefly sum up our contributions as follows:
\begin{itemize}
    \item We propose~\datasetname{},  a large-scale synthetic dataset with diversity in objects and their relative position and orientation in the scene.
    \item Further, we create a pipeline to add multiple objects to a scene in~\datasetname{}.
    \item We show that with a curriculum strategy of training on~\datasetname{}, a generative method can also generalize to real-world scenes. We show this generalization capability on the challenging real-world MSD~\cite{Yang_2019_ICCV} dataset.
\end{itemize}

\begin{table}[!t]
 \caption{
 Our proposed dataset,\textbf{~\datasetname{}}, surpasses existing mirror datasets in terms of attribute diversity and variability. While recent work~\cite{reflectingreality} introduced the synthetic~\olddataset{} dataset, it lacks key augmentations and scenario, limiting its performance in complex and real-world settings (See~\cref{fig:teaser}).
 }
\begin{adjustbox}{width=\linewidth}
\begin{tabular}{@{}l|cccc@{}}
\toprule
\multicolumn{1}{c|}{\textbf{Dataset}} & \textbf{Type} &  \textbf{Size ($\#$Images)} & \textbf{Attributes}\\ \midrule
 MSD~\cite{Yang_2019_ICCV} & Real  & 4,018 &  RGB, Masks \\
 Mirror-NeRF~\cite{zeng2023mirror-nerf} & Real \& Synthetic  & 9 scenes &  RGB, Masks, Multi-View\\
 DLSU-OMRS~\cite{DLSU} & Real   & 454 &   RGB, Mask \\
 TROSD~\cite{sun2023trosd} & Real  & 11,060  & RGB, Mask\\
 PMD~\cite{PMD:2020} & Real   & 6,461 &   RGB, Masks\\
 RGBD-Mirror~\cite{mei2021depth} & Real   & 3,049 &   RGB, Depth\\
 Mirror3D~\cite{mirror3d2021tan} & Real   & 7,011 &  RGB, Masks, Depth\\ 
 ~\olddataset{}~\cite{reflectingreality} & Synthetic   & 198,204 &  \makecell{Single Fixed Objects: RGB, Depth,\\ Masks, Normals, Multi-View}\\ 
 \midrule
\textbf{\datasetname(Ours)} & \makecell{Synthetic with Single \& \\ Multiple Objects} & \textbf{207,610} &  \makecell{Single + \textbf{Multiple Objects}: RGB, Depth,\\ Masks, Normals, Multi-View, \textbf{Augmentations} }\\ \bottomrule
\end{tabular}
\end{adjustbox}
\label{tab:data_characteristic}
\end{table}

%% file: sec/2_related_work.tex
\section{Related Work}
\label{sec:related_work}

\noindent
\textbf{Image Generative models.} 
Diffusion models~\cite{sohl2015deep} have become quite popular for image generation tasks. Diffusion models work by gradually adding noise to data and then learning to reverse this process to generate data from a variety of distributions~\cite{daras2022soft,dhariwal2021diffusion, ho2020denoising}. Subsequent works have expanded the scope of image generation by incorporating text guidance~\cite{ramesh2022hierarchical,saharia2022photorealistic} into the diffusion process, simplifying the reverse process~\cite{wallace2023edict}, and reformulating diffusion to occur in a latent space~\cite{rombach2022high} for improved speed.~\cite{parihar2024balancing} explore advancements in diffusion models by addressing bias through distribution-guided debiasing techniques. Further, methods~\cite{parihar2024text2place,parihar2024precisecontrol} are developed to provide more fine-grained generation control to these models.  
Building on the success of vision transformers~\cite{vaswani2017attention}, recent approaches~\cite{peebles2023scalable,Zheng2023FastTO,crowson2024scalable} have replaced the U-Net architecture in diffusion models with transformer-based designs, leading to high-quality image generation results. Further, there are popular methods~\cite{flux,sd35} for high-quality image generation. However, these methods also fail for the task of generating reflections on the mirror as shown in~\cref{fig:motivation}

\noindent
\textbf{Image Inpainting.} 
Building on the advancements in image diffusion models, methods like Palette~\cite{saharia2022palette} and Repaint~\cite{lugmayr2022repaint} leverage known regions through the denoising process to reconstruct missing parts. Blended Diffusion~\cite{avrahami2022blended, avrahami2023blended} refines this approach by replacing noise in unmasked areas with known content but struggles with complex scenes and shapes. Stable-Diffusion Inpainting~\cite{rombach2022high} (SDI) enhances results by fine-tuning the denoiser with noisy latents, masks, and masked images. Recent methods, such as HD-Painter~\cite{manukyan2023hd}, PowerPaint~\cite{zhuang2023task}, SmartBrush~\cite{xie2023smartbrush} build on SDI with additional training. Recently, BrushNet~\cite{ju2024brushnet} introduces a plug-and-play architecture that preserves unmasked content while improving coherence with textual prompts. 
However,~\cref{fig:teaser} highlights the limitations of these methods in generating reflections on the mirror.

\noindent
\textbf{Diffusion Models and 3D concepts.} Recently, LRM~\cite{lrm} based methods predict 3D model from a single image. Some methods~\cite{instructive3D} utilize diffusion-based methods to enable editing of these 3D presentations. Other diffusion-based methods~\cite{michel2024object} use synthetic image pairs for 3D-aware image editing. However, the synthetic-to-real domain gap can limit their applicability. Further, ObjectDrop~\cite{winter2024objectdrop} trains a diffusion model for object insertion/removal using a counterfactual dataset that can handle shadows and specular reflections.~\citet{sarkar2024shadows} shows that generated images have different geometric features such as shadows and reflections from the real images.~\citet{upadhyay2023enhancing} proposed a geometric constraint in the training process to improve the perspective cues in the generated images. Alchemist~\cite{sharma2024alchemist} provides control over the material properties of an object by proposing an object-centric synthetic dataset with physically-based materials. 

%% file: sec/3_method.tex
\section{Dataset}
\label{sec:dataset}

\begin{figure*}
    \centering
    \includegraphics[width=\linewidth]{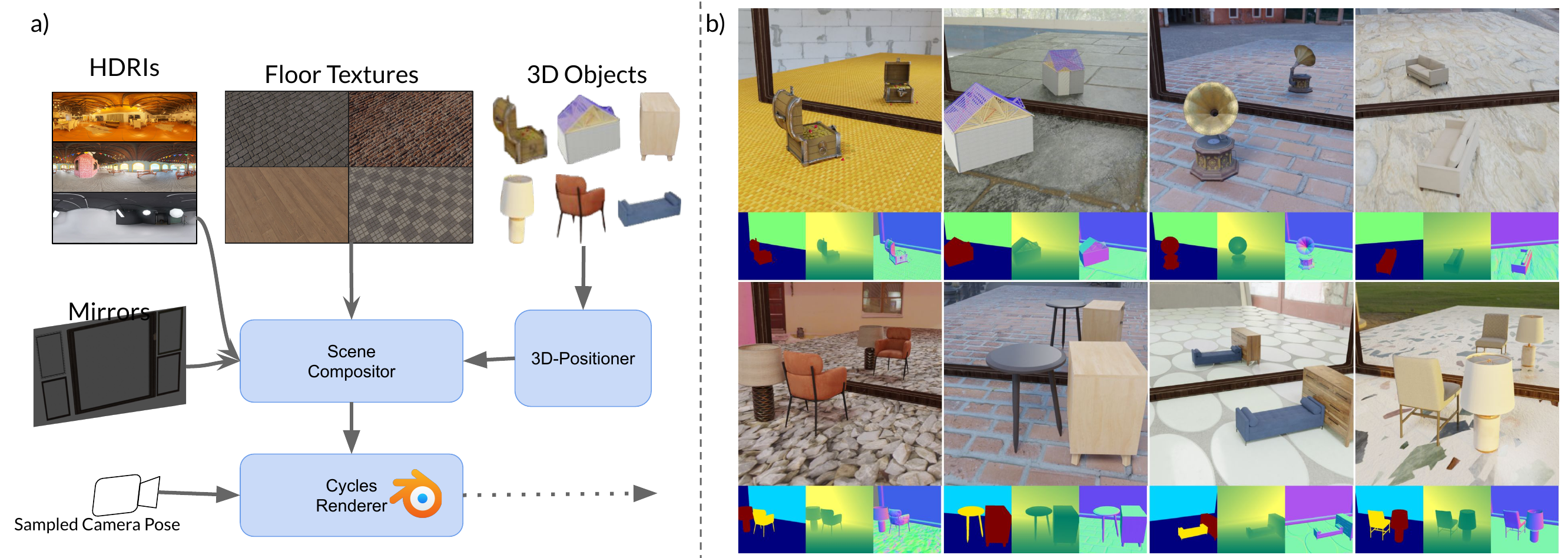}
    \caption{
\textbf{Dataset Generation Pipeline.} Our dataset generation pipeline introduces key augmentations such as \textbf{random positioning, rotation, and grounding} of objects within the scene using the 3D-Positioner. Additionally, we pair objects in semantically consistent combinations to simulate complex spatial relationships and occlusions, capturing realistic interactions for \textbf{multi-object} scenes.
}

    \label{fig:dataset_generation}
\end{figure*}
%Para1: Primary Goal: Motivate, Introduce and Summarize. 1. Previous methods fail on this task. 2. Contemporary dataset is able to solve to some extent but fails on out-of-distribution dataset. 3. Introduce the dataset. 4. Summary of the objects, rendeings and new augmentations in this dataset. 5. Discuss the new benchmark as well here.

\subsection{Data Generation Pipeline}
\cref{fig:motivation} highlights the failure of state-of-the-art models in handling the reflection generation task. MirrorFusion~\cite{reflectingreality} addresses this challenge by proposing a synthetic dataset but struggles in complex scenarios involving multiple objects and real-world scenes (~\cref{fig:teaser}). We attribute this limitation to the lack of diversity in their dataset. To mitigate these shortcomings, we introduce~\datasetname{}, a large-scale dataset which significantly expands diversity with varied backgrounds, floor textures, objects, camera poses, mirror orientation, object positions, and rotations.~\cref{tab:data_characteristic} compares existing mirror datasets, while~\cref{fig:dataset_generation} showcases samples from~\datasetname{}. 

%Para2: 1. Introduce Object sources. 2. Describe challenges while sourcing these objects. 3. Refer how spurious objects affect and how they were solved. 4. Summarize the total objects used in the renderings. 
%Manan will ask that why GSO is mentioned here. Ans: We will use it in test-set only. 

\noindent
\textbf{Object Sources.} 
We source objects from Objaverse~\cite{deitke2023objaverse} and Amazon Berkeley Objects (ABO)~\cite{collins2022abo} datasets. Objaverse, a large-scale dataset, contains $800K$ diverse 3D assets, while ABO contributes $7,953$ common household objects. To ensure quality, we refine our selection using a curated list of $64K$ objects from OBJECT 3DIT~\cite{michel2024object} and the filtering procedure discussed in~\cite{reflectingreality}, eliminating low-quality textures and sub-par renderings. After filtering, we get $58,109$ objects from Objaverse. In total, we utilize $66,062$ objects.

%Para3: 1. Introuce scene settings. 2. 2. Refer that floor and HDRI are randomized. 
\noindent
\textbf{Scene Resources.} To create a realistic scene, we require assets such as a mirror, floor and background. We create a plane for the floor and apply diverse textures sourced from CC-textures~\cite{denninger2023blenderproc2}.
We use HDRI samples provided by PolyHaven~\cite{polyhaven} to represent the background. In our experiments, we use different kinds of mirrors: full-wall mirrors and tall rectangular mirrors. For lighting, we position an area-light slightly above and behind the object at a $45^{\circ}$ angle, directing it towards both the object and the mirror.

%Para4. Come to objects and Describe the augmentations for placement of the objects.
\noindent
\textbf{Object Placement in the scene.} To begin, we fix the mirror’s position within the scene as a fixed reference point. The sampled object is then scaled to fit within a unit cube, ensuring uniformity in size across all objects. We proceed by sampling the object’s x-y position from a pre-computed region that guarantees both visibility of the object in the mirror and camera. This pre-computed region is determined by identifying the intersection between the mirror’s viewing frustum and the camera’s viewing frustum. Once the position is set, we randomly sample an angle for the object’s rotation around the y-axis to introduce variability. However, even with these steps, there may be instances where the object appears to float in the air, which can undermine the dataset’s utility. To address this, we apply a straightforward grounding technique, detailed in the supplementary material. Together, these strategies contribute to the diversity and overall quality of the proposed dataset.

%ALGORITHM FOR MULTIPLE OBJECTS
\begin{algorithm}[!t]
\caption{Procedure to Render Multiple Objects}
\begin{algorithmic}[1]
\REQUIRE Input 3D model $\mathcal{M}_1$
%\ENSURE Output result $R$

\STATE \textbf{Function} \textsc{GetPairedObjectCategory}($M$)
\STATE $c \gets \text{GetSemanticCategory}(M)$
\STATE $L \gets \text{GetPairedCategoriesList}(c)$
\STATE $c_{paired} \gets \text{SampleCategory}(L)$
\RETURN $c_{paired}$

\STATE \textbf{Function} \textsc{SampleObject}($c$)
\STATE $L_{obj} \gets \text{GetListObjects}(c)$
\STATE $\mathcal{M} \gets \text{Sample3DObject}(L_{obj})$
\RETURN $\mathcal{M}$

\STATE \textbf{Main Algorithm}
\STATE $c_{paired} \gets \textsc{GetPairedObjectCategory}(\mathcal{M}_1)$
\STATE $\mathcal{M}_2 \gets \textsc{SampleObject}(c_{paired})$
\STATE Initialize position of $\mathcal{M}_2$ at $X$
\WHILE{$\mathcal{M}_2$ collides with $\mathcal{M}_1$}
    \STATE $T_r \gets \text{SampleRandomPosition}()$
    \STATE $X \gets T_r$
\ENDWHILE

%\STATE \textbf{Function} \textsc{SampleObject}($c$)
\end{algorithmic}
\label{algo:multiple}
\end{algorithm}

\noindent
\textbf{Multiple Objects.}
A typical scene includes multiple objects arranged in varied layouts, producing a range of depth and occlusion scenarios that enhance scene realism. To capture this complexity, our dataset incorporates scenes with multiple objects, as described in~\cref{algo:multiple}. We start by sampling \( K \) objects from the original ABO dataset and identifying each object's class from~\cite{collins2022abo}. Categories are manually paired to ensure semantic coherence—for instance, pairing a chair with a table. During rendering, after positioning and rotating the primary object \( K_1 \), an additional object \( K_2 \) from the paired category is sampled and arranged to prevent overlap, ensuring distinct spatial regions within the scene. This process yields 3,140 scenes featuring diverse object configurations and spatial relationships, providing a robust foundation for realistic scene representation.

%Para5: 1. Describe settings for rendering. 2. Number of camera poses. 3. Blender cycles etc. 
\noindent
\textbf{Rendering.} 
Following scene composition, we randomly sample three camera poses from a predefined list of $19$ camera positions and render each scene using BlenderProc~\cite{denninger2023blenderproc2} to obtain RGB, depth, normal, and semantic label outputs. All renderings are produced at a resolution of $512 \times 512$ pixels. We set the ``cycles rendering'' parameter to $1024$, which is necessary for accurately capturing reflections. Representative samples are provided in~\cref{fig:dataset_generation} and additional examples are available in the supplementary material.

\section{Method}
\label{sec:model_training_details}

\paragraph{Preliminaries}
Diffusion models are generative models that can construct data samples by progressively removing noise. In the forward diffusion process, Gaussian noise $\epsilon \sim \mathcal{N} \left( 0,1 \right)$ is incrementally added to an initial clean sample $x_0$ over  $T$  timesteps to create a noisy sample $x_T$. In the reverse process, a clean image $x_0$ is reconstructed by iteratively denoising $x_T$. This denoising process is carried out by a denoising network $\epsilon_{\theta}$ which is conditioned on the timestep $t\in\{1,T\}$ and optional additional conditioning $c$ (e.g. text prompts, inpainting masks). Training loss of the denoiser is as follows:
\begin{equation}
L_{DM} = \;\; E_{x_0,\epsilon \sim \mathcal{N} \left( 0,I \right) ,t} ||\epsilon -\epsilon _{\theta}\left(z_t, t, c\right)||^2 
 \label{eq:train_objective}
\end{equation}

\begin{figure}[!t]
    \centering
    \includegraphics[width=\linewidth]{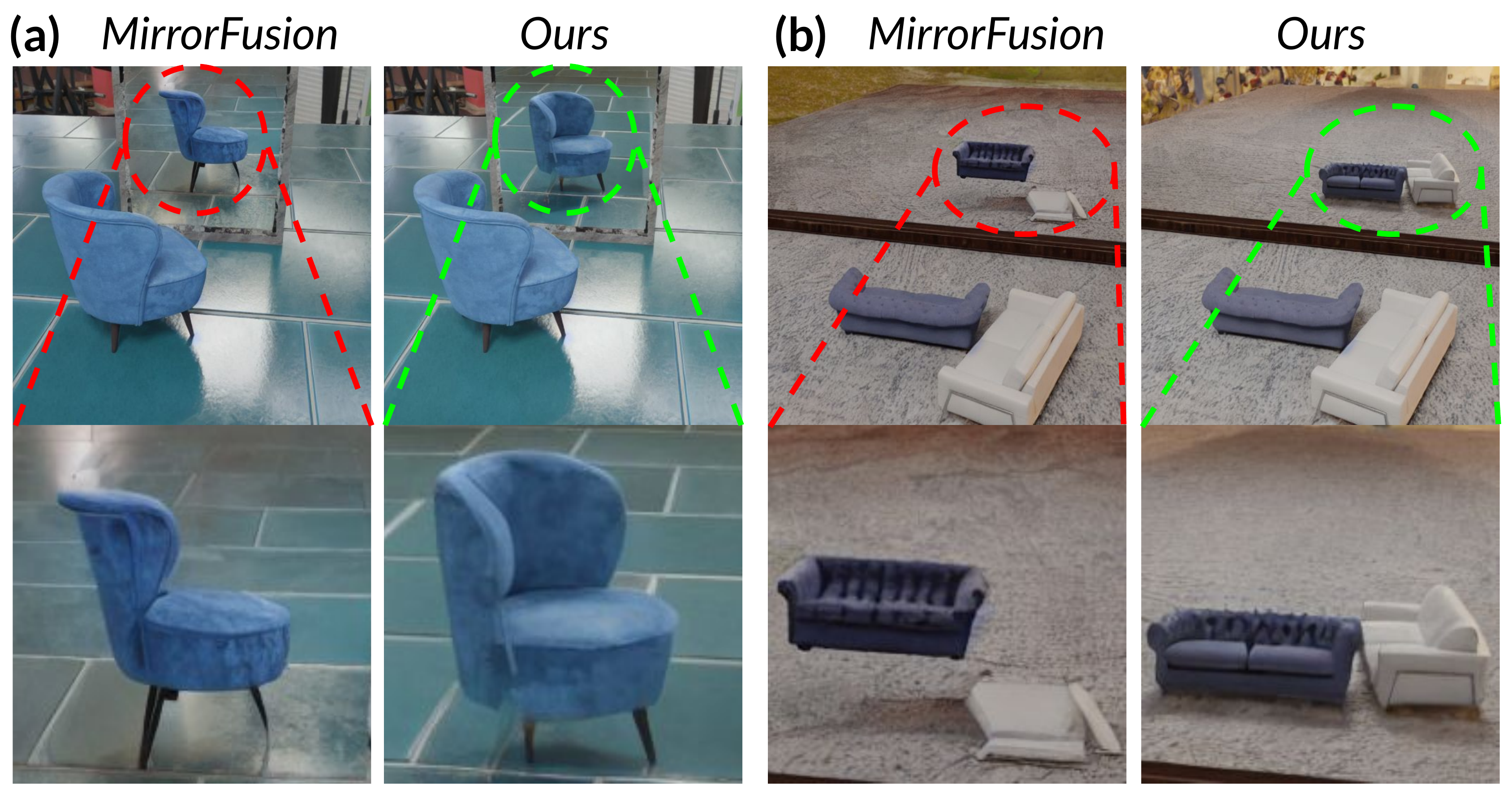}
    \caption{\textbf{Comparison on~\testsetname{}.}
The baseline fails to maintain accurate reflections and spatial consistency, showing (a) incorrect chair orientation and (b) distorted reflections of multiple objects. In contrast, our method correctly renders (a) the chair and (b) the sofas with accurate position, orientation, and structure, demonstrating superior performance.}

    \label{fig:highlight_issues}
\end{figure}

\paragraph{Model Architecture.}

Building upon~\oldmethod{}~\cite{reflectingreality}, we also formulate this task as an inpainting task. Our model employs a base dual branch network similar to  BrushNet~\cite{ju2024brushnet} and additionally uses depth map conditioning for the condition branch of BrushNet.
In particular, we concatenate the noisy latent $z_t$, masked image $z_m$, inpainting mask $x_m$ and depth map $x_d$, and provide this as an input to the conditioning U-Net
% ${{}^{BN}\epsilon^{'}}$ 
branch. Each layer of the generation U-Net ${\epsilon_i}$ is conditioned with the corresponding layer of the conditioning U-Net ${\epsilon^{'}}$  with the help of zero-convolutions ($\mathcal{Z}$) as follows:
\vspace{-2mm} 
\begin{equation}
 \epsilon _{\theta}\left( z_t,t,c \right) _i={}\epsilon _{\theta}\left( z_t,t,c \right) _i+  
 w\cdot\mathcal{Z} \left( {}\epsilon _{\theta}^{'}\left( \left[ z_t,z_m,x_m,x_d \right] ,t \right) _i \right) 
 \label{eq:conditioning_eq_bn}
\end{equation}

\noindent $w$ is the preservation scale to adjust the influence of conditioning. We set $w$ to be $1.0$ for all our experiments. We, train the model with the loss in~\cref{eq:train_objective}.

\noindent
\textbf{Training details.}
We follow a $3$ stage training curriculum to improve the generalization of the model on real-world scenes. We utilize the AdamW~\cite{adamw} optimizer with a learning rate of $1e^{-5}$ and a batch size of $4$ per GPU. We train on $4$ NVIDIA A100 GPUs in all stages.
\begin{itemize}

    \item \textbf{\textit{Stage 1.}} 
    In the first stage, we initialize the weights of both the conditioning and generation branch with the Stable Diffusion v1.5 checkpoint and finetune the model on the single object train split of our proposed~\datasetname{}. In contrast to~\cite{reflectingreality}, we do not keep the generation branch frozen and train the model till $40,000$ iterations. The variation in the position and rotation in the~\datasetname{} compared to~\olddataset{} allows us to train the model for longer iterations without any degradation in the generation quality compared to~\cite{reflectingreality}.

    \item \textbf{\textit{Stage 2.}} In the second stage, we finetune the model for $10,000$ iterations on the multiple objects train split of~\datasetname{} to incorporate the concepts of occlusions as present in realistic scenes.

    \item \textbf{\textit{Stage 3.}} We propose a third stage training on real-world data from the MSD~\cite{Yang_2019_ICCV} dataset for another $10,000$ iterations to bridge the domain gap between synthetic and real-world image inpainting.
\end{itemize}

In the first two stages, we use ground truth depth maps and for the third stage, we generate depth maps using a monocular depth estimator~\cite{bochkovskii2024depthpro}. To enhance learning and reduce reliance on text prompts, we randomly drop them $20\%$ of the time during training, enabling the model to utilize depth information better.

\noindent
\textbf{Inference.} During inference, we use a CFG value of $7.5$ and utilize the UniPC scheduler~\cite{zhao2024unipc} for 50 time steps. During inference, we allow the user to provide the mask depicting the mirror and estimate the input depth map using Depth-Pro~\cite{bochkovskii2024depthpro} by passing the masked image as input.

\begin{figure}[!t]
    \centering
 \includegraphics[width=\linewidth]{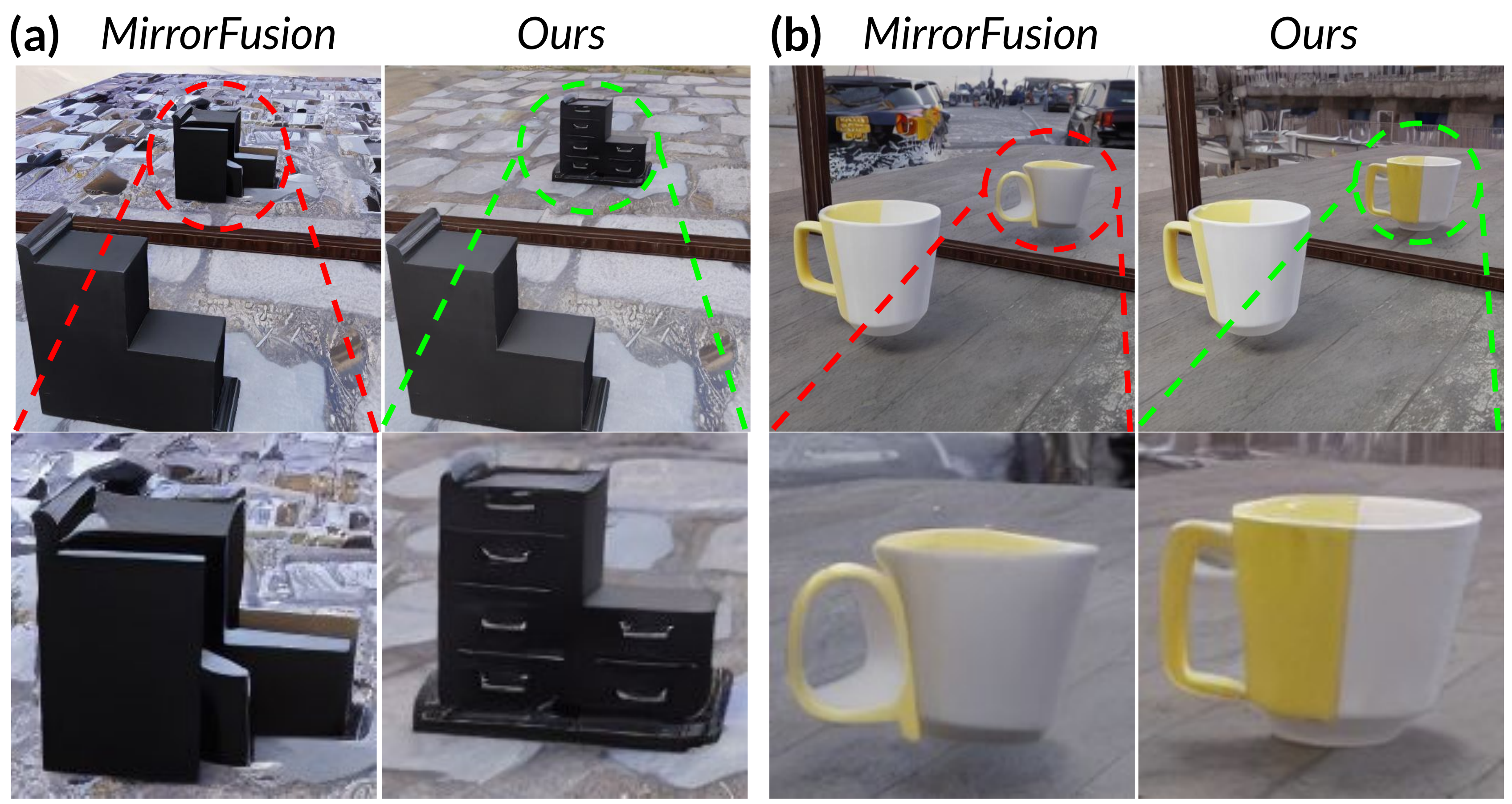}
    \caption{\textbf{Comparison on GSO~\cite{downs2022google} dataset.} In (a), the baseline method misrepresents object structure, while our method preserves spatial integrity and produces realistic reflections. In (b), the baseline yields incomplete and distorted reflections of the mug, whereas our approach generates accurate geometry, color, and detail, showing superior performance on out-of-distribution objects.}

    \label{fig:gso}
\end{figure}

%% file: sec/4_results.tex
\section{Experiments \& Results}
\label{sec:results}
% Para1: Introduction of baselines

\noindent
We discuss the evaluation strategy and compare our current method with the previous state-of-the-art method, ~\oldmethod{}~\cite{reflectingreality}, referring to this as the baseline. Additionally, we also provide ablation studies on different design choices in~\cref{sec:ablation_studies}. 

% Para2: Dataset for benchmarking
\vspace{1mm}
\noindent 
\textbf{Dataset.} Compared to~\oldtestset{},~\testsetname{} consists of renderings of single and multiple objects in a scene. Additionally, we qualitatively test our method on several images from the MSD dataset and renderings from the Google Scanned Objects(GSO)~\cite{downs2022google} dataset. For single object renderings, we have a total of $2,991$ images, which come from categories that are both seen and unseen during training. We create 300 images that contain two objects from the ABO dataset in the same scene to test the model on generating reflections for multiple objects.

\begin{figure*}
    \centering
    \includegraphics[width=\linewidth]{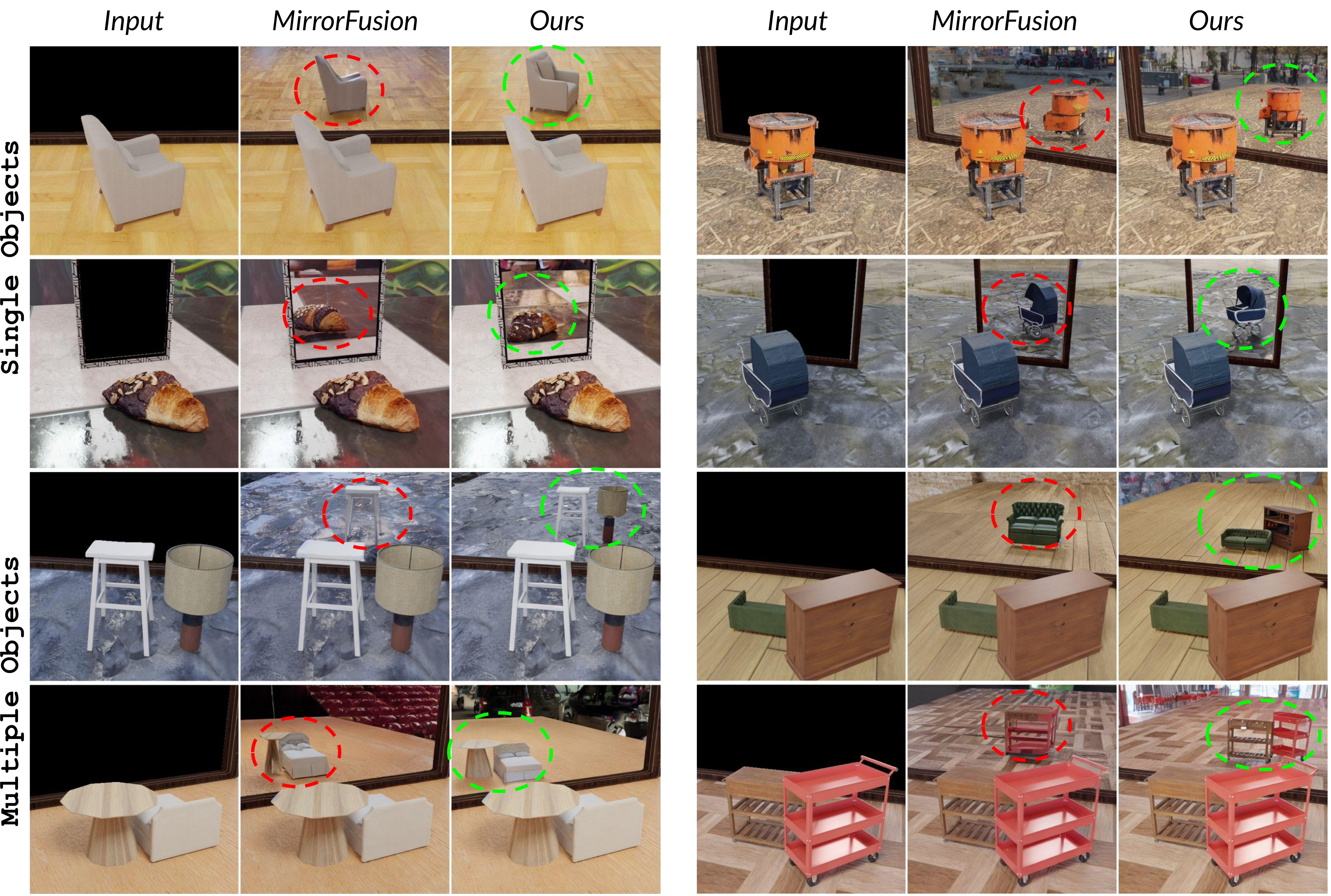}
    \caption{\textbf{Results on~\testsetname{}.} We compare our method with the baseline~\oldmethod{}~\cite{reflectingreality} on~\testsetname{}. The baseline method shown struggles with pose variations, even in single-object scenes, and fails to produce accurate reflections for multiple objects. In contrast, our method handles variations in the object orientation effectively and generates geometrically accurate reflections, even in complex, multi-object scenarios.}
    \label{fig:qual_big}
\end{figure*}

% Para3: Metrics
\noindent
\textbf{Metrics.}
We benchmark various methods on the
% preservation of the unmasked region, 
quality of the generated reflection and textual alignment of the generated image with the input prompt.

\begin{itemize}
 
    \item \textbf{\textit{Reflection Generation Quality.}} We evaluate reflection quality using Peak-Signal-to-Noise ratio (PSNR), Structural Similarity (SSIM) and Learned Perceptual Image Patch Similarity (LPIPS)~\cite{zhang2018unreasonable} on the masked mirror region.

    \item \textbf{\textit{Text Alignment.}} We use CLIP~\cite{radford2021learning} Similarity for assessing textual alignment. 
\end{itemize}

\noindent
\textbf{Qualitative results on~\testsetname{}.} 
 In~\cref{fig:highlight_issues} (a), a single chair that is slightly rotated is placed in front of a mirror. We observe that the baseline method completely misrepresents the chair's orientation in the generated reflection as seen in the mirror. Notice the zoomed-in region where the reflection appears as if the object was cut and pasted onto the mirror.  In contrast,~\methodname{} trained on~\datasetname{} accurately captures the chair's orientation in the reflection, as shown in the zoomed-in region highlighted by the green circle.

~\cref{fig:highlight_issues} (b), shows a scene with a white sofa rotated and placed to the right of a gray sofa. The baseline method produces two artifacts in the reflection: 1) the gray sofa appears to be floating in the air, and 2) the generated reflection of the white sofa is completely incorrect. In contrast, our method accurately generates the scene in the reflection. These results demonstrate the effectiveness of our augmentation strategies, as described in~\cref{sec:dataset}. We show more examples with both single and multiple objects in~\cref{fig:qual_big}.

\noindent
\textbf{Qualitative results on GSO~\cite{downs2022google}.}
We further evaluate the generalization ability of~\methodname{} on real-world scanned objects from GSO, shown in~\cref{fig:gso}.~\methodname{} generates significantly more accurate and realistic reflections. For instance, in~\cref{fig:gso} (a),~\methodname{} correctly reflects the drawer handles (highlighted in green), while the baseline model produces an implausible reflection (highlighted in red). Likewise, for the ``White-Yellow mug'' in~\cref{fig:gso} (b),~\methodname{} delivers a convincing geometry with minimal artifacts, unlike the baseline, which fails to accurately capture the object's geometry and appearance.

\begin{figure*}[!t]
    \centering
    \includegraphics[width=\linewidth]{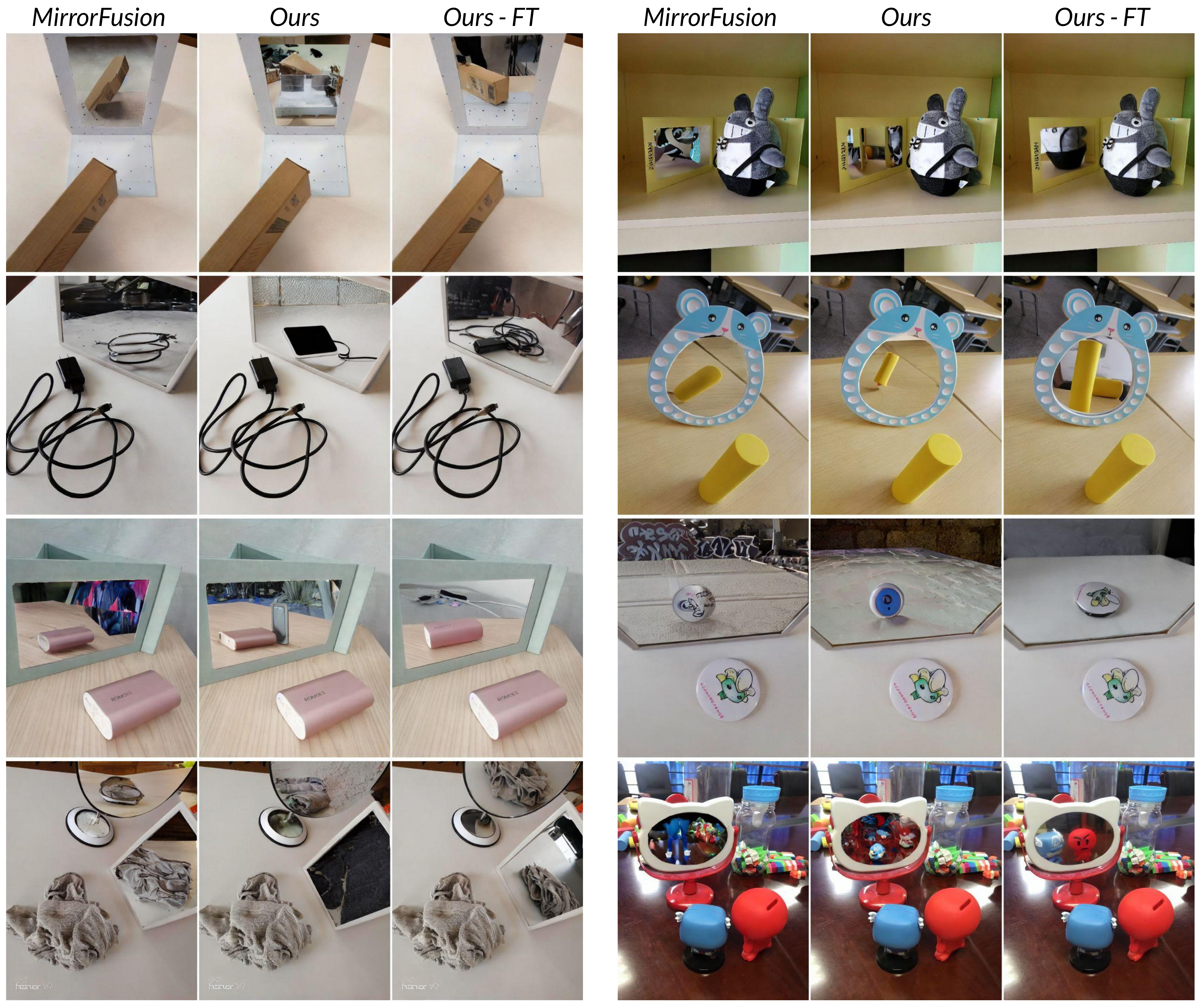}
    \caption{\textbf{Real World Scenes.} We show results for MirrorFusion~\cite{reflectingreality}, our method and our method fine-tuned on the MSD~\cite{Yang_2019_ICCV} dataset. We observe that our method can generate reflections capturing the intricacies of complex scenes, such as a cluttered cable on the table and the presence of two mirrors in a 3D scene.}
    \label{fig:msd_real_world}
    \vspace{-2mm}
\end{figure*}

\noindent
\textbf{Qualitative results on the Real-World MSD dataset.}
~\methodname{} performs well on~\testsetname{} and real-world objects from GSO but struggles with complex scenes, such as cluttered cables on a table and reflections across multiple mirrors (see~\cref{fig:msd_real_world}). To improve coherence, we fine-tune it on a subset of the MSD dataset and test it on a held-out split, enhancing its ability to handle real-world scenarios. As shown in~\cref{fig:msd_real_world}, this fine-tuning enables high-fidelity reflections, accurately capturing details like the ``black cable'' on the table and the ``towel'' in both mirrors. These results demonstrate how our dataset improves diffusion models, enabling more realistic reflections in challenging settings.~\cref{fig:msd_real_world} illustrates further examples on the real-world MSD dataset.

\noindent
\textbf{Quantitative results with baselines.}
For evaluating the metrics, we generate images using four seeds for a particular prompt and select the image that has the best SSIM score on the unmasked region. For a particular metric, we report the average value across~\testsetname{} by averaging the metric for all the selected images.~\cref{tab:single_object,tab:multiple_object} show that our method outperforms the baseline method and finetuning on multiple objects improves the results on complex scenes. 

\begin{table}[!t]
\centering
\caption{\textbf{Single Object Reflection Generation Quality.} We compare the quantitative results between the baseline and~\methodname{} on the \textbf{single object} split of~\testsetname{}. The best results are shown in \textbf{bold}. This shows the effectiveness of the dataset by achieving improved scores.}
\label{tab:single_object}
\begin{adjustbox}{width=\linewidth}
\begin{tabular}{@{}c|ccc|c@{}}
\toprule
Metrics & \multicolumn{3}{c|}{Reflection Generation Quality} & Text Alignment \\ \midrule
Models & \textbf{PSNR} $\uparrow$ & \textbf{SSIM} $\uparrow$ & \textbf{LPIPS} $\downarrow$ & \textbf{CLIP Sim} $\uparrow$ \\ \midrule
\textbf{baseline}~\cite{reflectingreality} & 18.31 & 0.76 & 0.122 & \textbf{26.00} \\
\textbf{Ours 40k} & \textbf{18.79} & \textbf{0.77} & \textbf{0.108} & 25.96 \\ \bottomrule
\end{tabular}
\end{adjustbox}
\end{table}

\begin{table}[!t]
\centering
\caption{\textbf{Multiple Object Reflection Generation Quality.} We compare the quantitative results between~\methodname{} trained without multiple objects and~\methodname{} trained with multiple objects on the \textbf{multiple object} split of~\testsetname{}. The best results are shown in \textbf{bold}. This shows the effectiveness of finetuning further on multiple objects.}
\label{tab:multiple_object}
\begin{adjustbox}{width=\linewidth}
\begin{tabular}{@{}c|ccc|c@{}}
\toprule
Metrics & \multicolumn{3}{c|}{Reflection Generation Quality} & Text Alignment \\ \midrule
Models & \textbf{PSNR} $\uparrow$ & \textbf{SSIM} $\uparrow$ & \textbf{LPIPS} $\downarrow$ & \textbf{CLIP Sim} $\uparrow$ \\ \midrule
\textbf{Ours 40k} & 17.77 & 0.743 & 0.126 & \textbf{26.17} \\
\textbf{Ours 50k} & \textbf{18.00} & \textbf{0.744} & \textbf{0.119} & 26.09 \\ \bottomrule
\end{tabular}
\end{adjustbox}
\end{table}

\noindent
\textbf{User study.} 
To evaluate the effectiveness of our proposed strategy, we also conducted a user study where we provided users with $40$ different samples containing single, multiple, GSO objects, and real-world generations from the baseline and~\methodname{}. \textbf{$84\%$  of users preferred generations from~\methodname{} over the baseline method}. We provide more details in~\cref{sec:supp_user_study}.

\begin{figure}
    \centering
    \includegraphics[width=\linewidth]{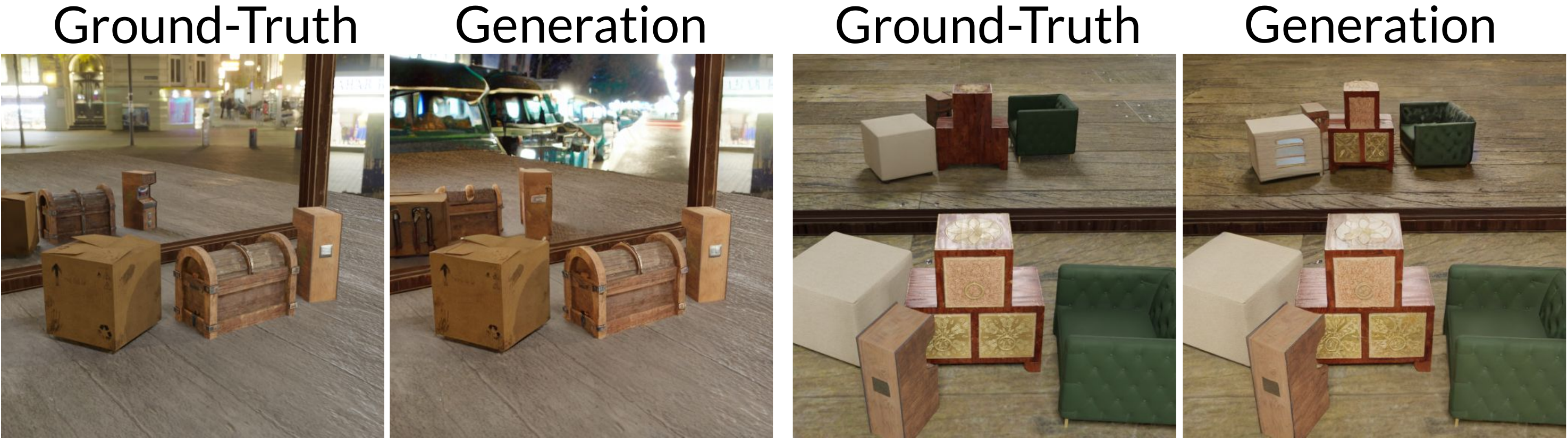}
\caption{\textbf{Limitations.} Our method performs well in multi-object scenes (more than two objects) but retains some artifacts, which can be reduced by synthesizing the dataset through the proposed data-generation pipeline and further increasing the diversity and scale.}    \label{fig:limitation}
\end{figure}

\noindent
\textbf{Limitations.} 
\cref{fig:limitation} illustrates examples where our method accurately captures overall geometry but introduces minor artifacts which can be easily addressed by synthesizing additional training data and fine-tuning the model.

\subsection{Ablation Studies}
\label{sec:ablation_studies}

\begin{figure}[!t]
    \centering
    \includegraphics[width=\linewidth]{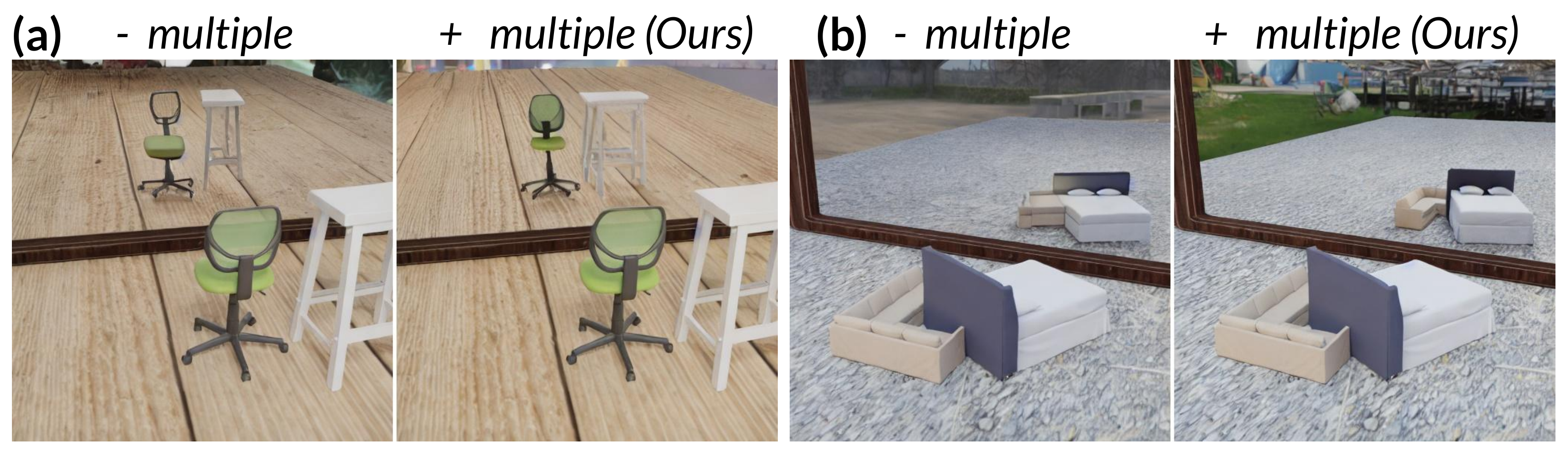}
\caption{\textbf{Impact of adding multiple objects.} We observe that training without multiple objects leads to (a) poor reflection generation and (b) artifacts like object blending, supporting the need for finetuning the model on such scenarios.}    
\label{fig:ablation_multiple}
\end{figure}

\noindent
\textbf{Impact of multiple objects dataset.}
To evaluate the impact of adding multiple objects to our dataset, we compare~\methodname{} with ( ``+ multiple'') and without (``- multiple'') object training in~\cref{fig:ablation_multiple}.``~\methodname{}-w/o multiple'' struggles to generate plausible mirror reflections, as evident in~\cref{fig:ablation_multiple} (b), where the bed and sofa appear to blend together. In contrast,``~\methodname{}-with multiple'' accurately captures the spatial relationships between objects within the mirror reflection. These results highlight the importance of including multiple objects in the dataset, enabling the model to learn spatial relationships and effectively handle occlusions.

\begin{figure}[!t]
    \centering
    \includegraphics[width=\linewidth]{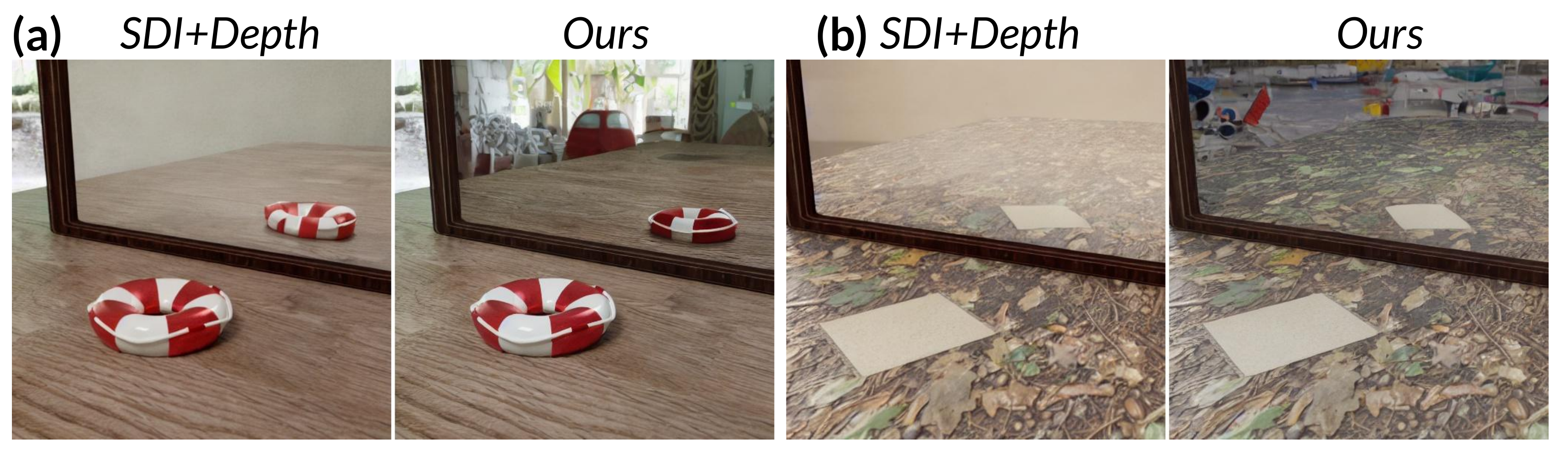}
\caption{\textbf{Comparison with SDI+Depth baseline.} We observe color leakage issues in ``SDI+Depth'' generations. A dual-branch architecture proves to be a better choice, yielding superior outcomes.}    \label{fig:ablation_architecture}
\end{figure}

\noindent
\textbf{Ablation on architecture.}
To further validate our architectural choice, we adapt Stable Diffusion Inpainting to accept depth maps as input similar to the changes made for~\methodname{} and train this modified model on our proposed dataset referring to it as ``SDI+Depth''. We compare ``SDI+Depth'' with~\methodname{} in~\cref{fig:ablation_architecture}. While ``SDI+Depth'' accurately positions objects in the mirror, it suffers from significant artifacts, including color leakage in contrast to~\methodname{}. We suspect that this happens due to the early combination of the noisy latent features, mask, and conditioning information in the initial convolution layer, restricting later layers from accessing clean features. These findings suggest that a dual branch architecture to provide the conditioning information separately as done in~\methodname{} is a better choice.

%% file: sec/5_conclusion.tex
\section{Conclusion}
We introduce \datasetname{}, a novel large-scale synthetic dataset designed to advance mirror reflection generation significantly. By employing targeted data augmentations, we achieved robust variability in object pose, position, and occlusion, alongside the ability to handle multi-object scenes. Our qualitative and quantitative evaluations demonstrate \datasetname{}'s efficacy in reflection generation, with promising generalization to real-world scenes using curriculum training. This dataset holds substantial potential for driving progress in various mirror-related tasks. Future research will explore advanced data augmentation techniques to enhance real-world performance further.

\noindent \textbf{Acknowledgment.} We are grateful to the Kotak IISc AI-ML Center for providing the computing resources. We thank Rishubh Parihar for their valuable feedback and Lokesh R. Boregowda for their valuable guidance.

%% file: sec/X_suppl.tex
\clearpage
\maketitlesupplementary

\appendix

\tableofcontents
\addtocontents{toc}{\protect\setcounter{tocdepth}{2}}

\section{Dataset Generation}
\label{sec:supp_dataset_generation}

\begin{algorithm}[t]
\caption{Procedure to Ground an Object}
\begin{algorithmic}[1]
\REQUIRE Input 3D model $\mathcal{M}$, Ground-Level $\underline{Z}$

\STATE \textbf{Function} \textsc{GroundObject}($\mathcal{M}, \underline{Z}$)
\STATE $bbox \gets \textsc{GetBoundingBox3D}(\mathcal{M})$

\IF{$bbox.z > \underline{Z}$}
    \STATE $z \gets bbox.z - \underline{Z}$
    \STATE $\Delta t \gets [0,0,-z]$
    \STATE $\mathcal{M}.position \gets \mathcal{M}.position + \Delta t$
\ENDIF

%\STATE \textbf{Function} \textsc{SampleObject}($c$)
\end{algorithmic}
\label{algo:ground_objects}
\end{algorithm}

\paragraph{Grounding Object.} We describe the algorithm to ground an object in~\cref{algo:ground_objects}. If the minimum value of the bounding box along the z-dimension is greater than the ground level, we adjust the object's position to rest it on the ground. This adjustment enhances the photorealism of the generated dataset, ensuring objects appear naturally grounded in their environment.

% \begin{figure}[t]
%     \centering
%     \includegraphics[width=\linewidth]{Figures/supplementary/sampling_region_top_view.pdf}
%     \caption{\textbf{Illustration to determine Sampling Region.}}
%     \label{fig:supp_sampling_region}
% \end{figure}

\paragraph{Sampling Region} 
For random positions in front of the mirror, we first define a region in the x-y plane where the object and its reflection are visible in the camera view. To determine this region, we compute the intersection of the camera's viewing frustum on the x-y plane with the extent of the mirror. %, as illustrated in~\cref{fig:supp_sampling_region}. 
This process is repeated for all camera locations used in the dataset generation. The resulting sampling region, $\mathcal{S}$, ensures that any position within it allows the visibility of both the object and its reflection in the camera view.

\begin{algorithm}[t]
\caption{Procedure to Place an Object in a scene}
\begin{algorithmic}[1]
\REQUIRE Input 3D model $\mathcal{M}$, Ground-Level $\underline{Z}$, Sampling region $\mathcal{S}$

\STATE \textbf{Function} \textsc{NormalizeObject}($\mathcal{M}$)
\STATE $bbox \gets \textsc{GetBoundingBox3D}(\mathcal{M})$
\STATE $d_{max} \gets \text{GetMaxDimension}(bbox)$
\STATE $s \gets \frac{1}{d_{max}}$
\STATE $S_{matrix} \gets \text{GetScaleMatrix(s)}$
\STATE $\mathcal{M}.scale \gets S_{matrix}$
\STATE $bbox \gets \textsc{GetBoundingBox3D}(\mathcal{M})$
\RETURN $bbox$

\STATE

\STATE \textbf{Function} \textsc{SamplePosition}($\mathcal{M}, S$)
\STATE $t \gets \text{SamplePosition}(S)$
\STATE $\mathcal{M}.translation \gets t$
\STATE $bbox \gets \textsc{GetBoundingBox3D}(\mathcal{M})$
\RETURN $bbox$

\STATE 

\STATE \textbf{Function} \textsc{RandomRotation}($\mathcal{M}$)
\STATE $\theta \gets \text{RandomAngle}(S)$
\STATE $R_{matrix} \gets \text{GetRotationMatrix(}\theta\text{)}$
\STATE $\mathcal{M}.rotation \gets R_{matrix}$
\STATE $bbox \gets \textsc{GetBoundingBox3D}(\mathcal{M})$
\RETURN $bbox$

\STATE 

\STATE \textbf{Main Algorithm}
%\STATE \textbf{Function} \textsc{GroundObject}($\mathcal{M}$)
\STATE $bbox \gets \textsc{NormalizeObject}(\mathcal{M})$
\STATE $bbox \gets \textsc{SamplePosition}(\mathcal{M}, \mathcal{S})$
\STATE $bbox \gets \textsc{RandomRotation}(\mathcal{M})$

%\STATE \textbf{Function} \textsc{SampleObject}($c$)
\end{algorithmic}
\label{algo:placement_objects}
\end{algorithm}

\paragraph{Object Placement}
We provide the procedure to place an object in a scene in~\cref{algo:placement_objects}. First, we scale an object to fit inside a unit cube, ensuring uniformity across objects. Next, we randomly sample a position in front of the mirror and update the position of the object accordingly. Finally, we apply a random rotation around the vertical axis. These steps guarantee that the object and its reflection remain visible in the camera's field of view. Further, these steps introduce greater diversity to the dataset.  

\paragraph{Additional Samples.} We provide additional samples for scenes with single object in~\cref{fig:supp_single_object} and multiple objects in ~\cref{fig:supp_multiple_object,fig:supp_more_than_two_objs}. Further, we also provide additional samples for visualization from~\datasetname{} in folder \textit{``dataset-samples''} in the attached supplementary material.

\section{Architecture Details}
\begin{figure}[!t]
    \centering
    \includegraphics[width=\linewidth]{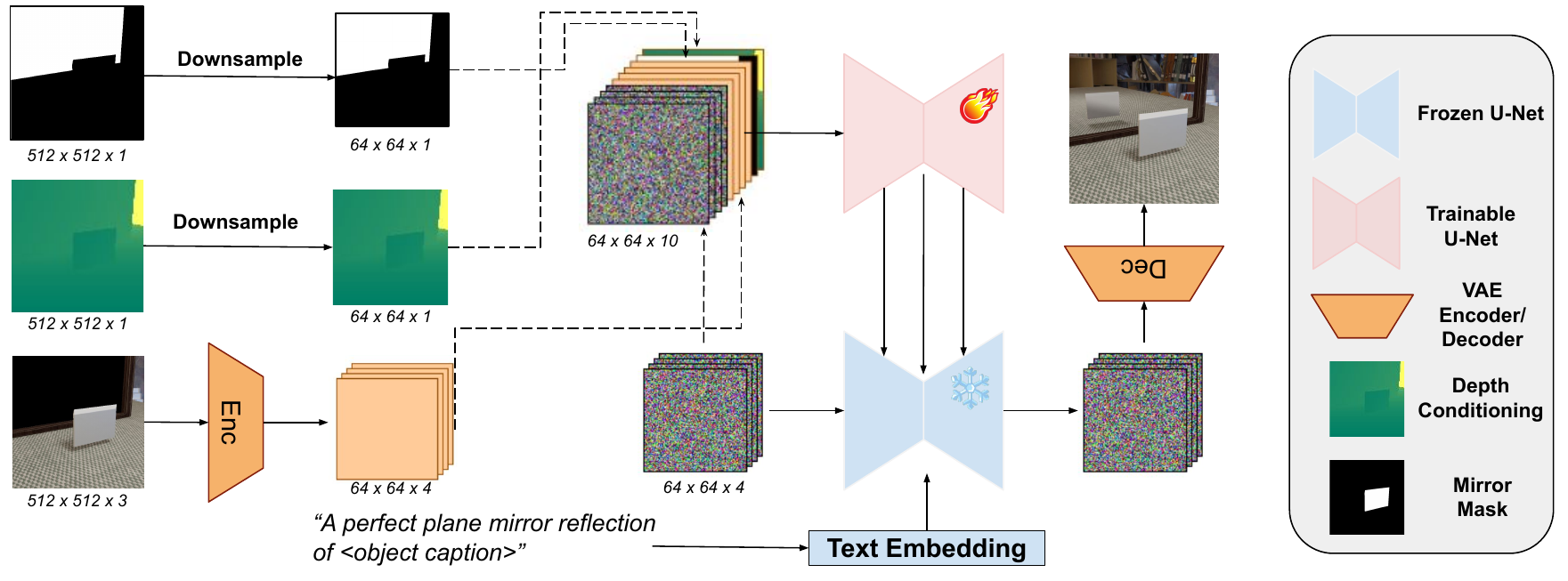}
    \caption{\textbf{Overview of architecture used for the experiments.}}
    \label{fig:supp_architecture}
\end{figure}
Our work builds upon the MirrorFusion framework ~\cite{reflectingreality}, which employs a conditioning network that leverages depth information to guide the generation process of a pre-trained generative network. This model is trained using a three-stage curriculum learning strategy, as detailed in~\cref{sec:model_training_details}.

\begin{figure}[!t]
    \centering
    \includegraphics[width=\linewidth]{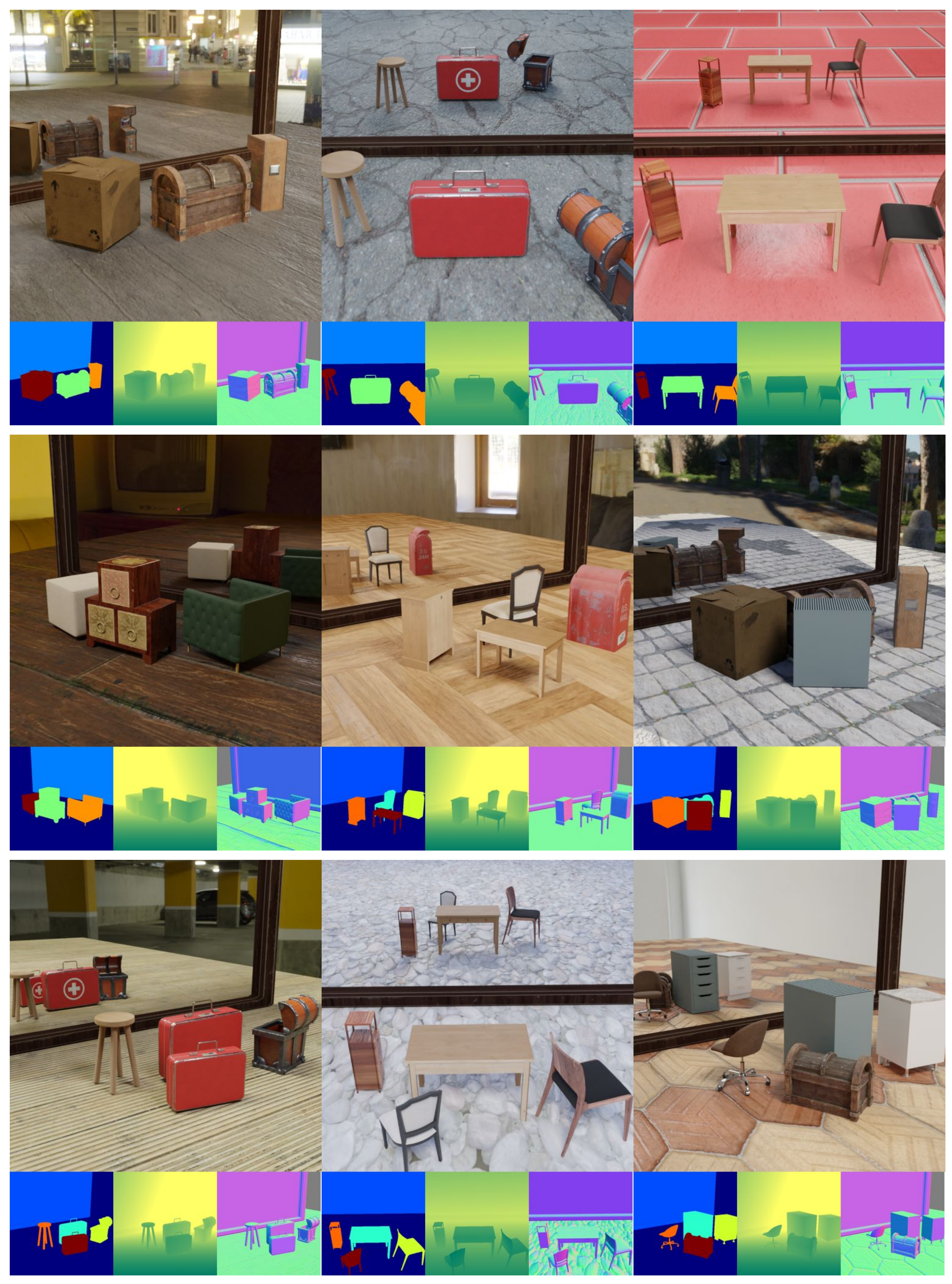}
    \caption{\textbf{Samples of scenes containing more than two objects.}}
    \label{fig:supp_more_than_two_objs}
\end{figure}

\begin{figure*}
    \centering
    \includegraphics[width=\linewidth, height=22cm, keepaspectratio]{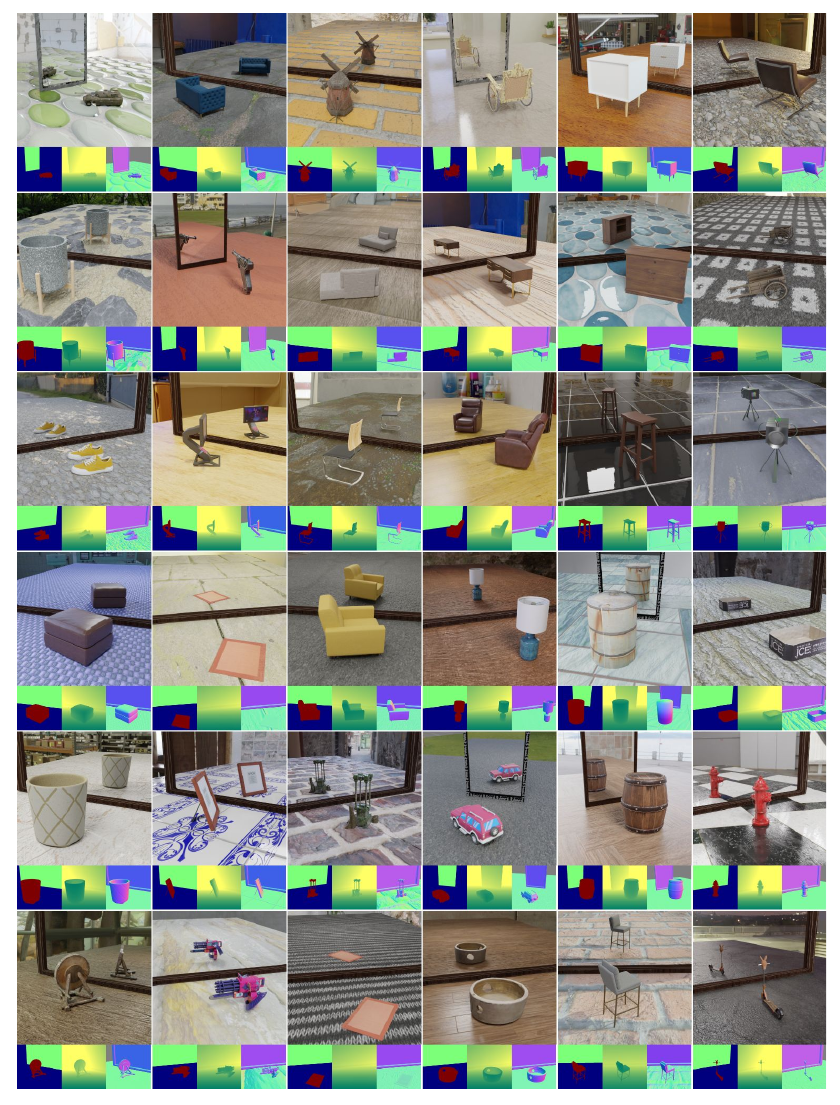}
    \caption{Samples of scene containing single object from~\datasetname{}}
    \label{fig:supp_single_object}
\end{figure*}

\begin{figure*}
    \centering
    \includegraphics[width=\linewidth,height=22cm, keepaspectratio]{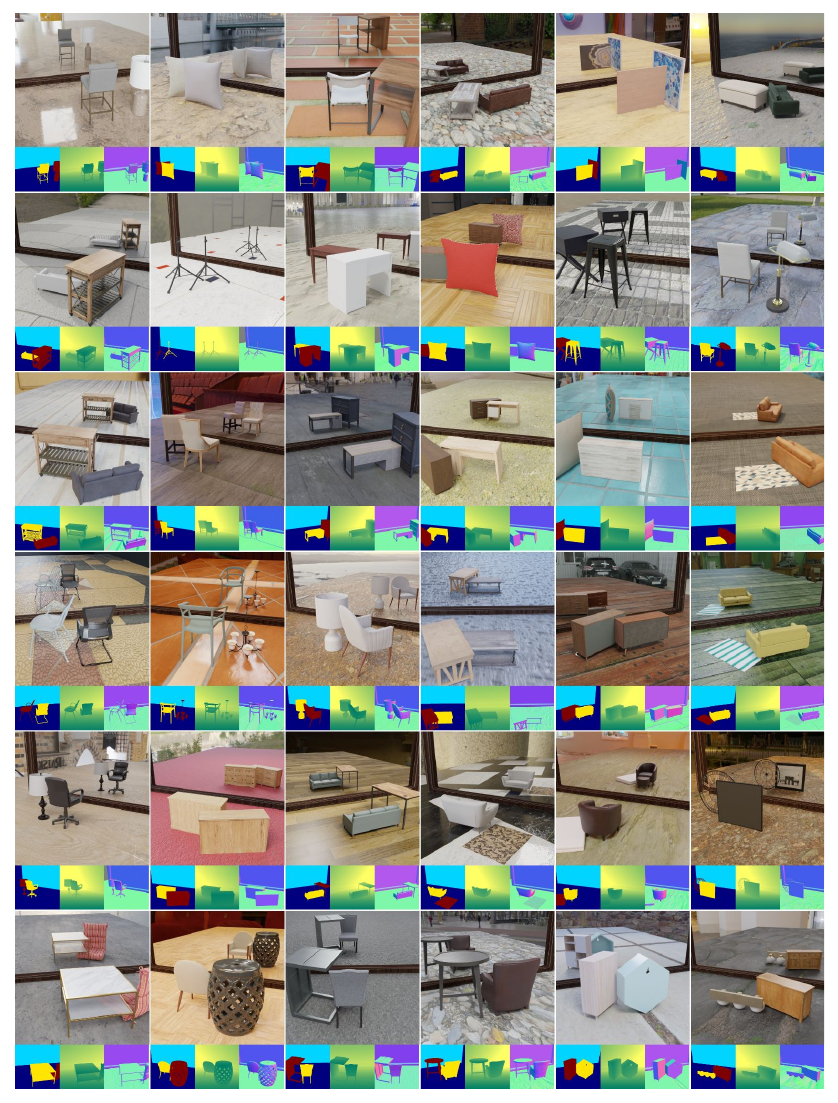}
    \caption{Samples of scene containing multiple objects from~\datasetname{}}
    \label{fig:supp_multiple_object}
\end{figure*}

\FloatBarrier

\section{Additional Details}
\label{sec:additional_details}

\subsection{Characteristic Comparison with~\olddataset{}}
\begin{table}[!t]
\caption{Comparison of the proposed dataset,~\datasetname{} , with~\olddataset{}}
\label{tab:supp_characteristic_table}
\begin{adjustbox}{width=\linewidth}
\begin{tabular}{@{}cccccc@{}}
\toprule
\textbf{Augmentations} & Grounding & \makecell{Random\\ Rotations} & \makecell{Random\\ Positions} & \makecell{Multiple\\ Objects} & \makecell{Occlusion\\ Scenarios} \\ \midrule
      \textbf{~\olddataset{}}        &      \xmark     &    \xmark              &    \xmark              &   \xmark               &   \xmark                  \\
       \textbf{~\datasetname{}}         &   \cmark        &    \cmark              &    \cmark              &   \cmark               &      \cmark               \\ \bottomrule
\end{tabular}
\end{adjustbox}
\end{table}
We present a characteristic comparison of the proposed dataset with~\olddataset{} in~\cref{tab:supp_characteristic_table}. This variety aids in the generalization of the~\methodname{} to complex scenarios and real-world scenes.

\subsection{Text prompts used in the experiments}
\label{sec:text_prompts_used}

We provide the text prompts used in the main paper for image generation.

\paragraph{Figure 1.} Each row in this figure uses the same text prompt. Text prompts are as follows:
\begin{itemize}
    \item \textbf{First row.} \textit{``A perfect plane mirror reflection of (A cylindrical bottle with a spherical top and bottom, featuring a neck and spout.) and (A cylinder with a conical bottom and a spherical top.)''}
    \item \textbf{Second row.} \textit{``A perfect plane mirror reflection of a yellow mug with a flower design is placed on a desk in front of a mirror. The reflection of the mug can be seen in the mirror, creating an interesting visual effect where the mug appears to be floating.''}
\end{itemize}

\paragraph{Figure 4.} Text prompts are as follows:
\begin{itemize}
    \item \textbf{(a)} \textit{``A perfect plane mirror reflection of a chair with a high, rounded back and blue upholstery.''}
    \item \textbf{(b)} \textit{``A perfect plane mirror reflection of (3D model of a Chesterfield sofa with cylindrical and spherical elements.) and (3D model of a two-seater sofa with backrest and armrests.)''}
\end{itemize}

\paragraph{Figure 5.} Text prompts are as follows:
\begin{itemize}
    \item \textbf{(a)} \textit{``A perfect plane mirror reflection of a black color furniture in stair shape with multiple drawers.''}
    \item \textbf{(b)} \textit{``A perfect plane mirror reflection of a yellow and white mug on a grey surface.''}
\end{itemize}

\paragraph{Figure 6.} Text prompts for ``Column-1'' are as follows:
\begin{itemize}
 \item \textbf{First row.} \textit{``A perfect plane mirror reflection of a curved and slatted 3D chair with backrest, seat, armrests, and footrest.''}
 \item \textbf{Second row.} \textit{``A perfect plane mirror reflection of a croissant that is chocolate and covered in nuts on one side and plain on the other.''}
 \item \textbf{Third row.} \textit{``A perfect plane mirror reflection of (A wooden stool with a backrest and a seat.) and (3D model of a lamp with a cylindrical body, spherical base, conical bottom, spherical top, and a cylindrical shade with a spherical accent.)''}
  \item \textbf{Fourth row.} \textit{``A perfect plane mirror reflection of (King size bed with a slatted base, tufted headboard and footboard, curved backrest and armrest, and slanted top and bottom edges.) and (A 3D object with a truncated octagonal base and a spherical top.)''}
\end{itemize}

Text prompts for ``Column-2'' are as follows:
\begin{itemize}
 \item \textbf{First row.} \textit{``A perfect plane mirror reflection of an orange vat sitting on a grey metal frame with a light gray colored control box attached.''}
 \item \textbf{Second row.} \textit{``A perfect plane mirror reflection of a dark blue baby buggy with no one in it.''}
 \item \textbf{Third row.} \textit{``A perfect plane mirror reflection of (A 3D model of a single-seater Chesterfield sofa with a tufted back, curved backrest, and slanted seat.) and (Rectangular cabinet with a slanted roof and base, featuring a door and a drawer, 3D modeled as a TV stand.)''}
  \item \textbf{Fourth row.} \textit{``A perfect plane mirror reflection of (3D model of a kitchen cart with three shelves, two drawers on each side, and a two-tiered bunk bed with slatted bases.) and (Three-tiered wheeled cart with shelves and a handle.)''}
\end{itemize}

\paragraph{Figure 7.} Text prompts for ``Column-1'' are as follows:
\begin{itemize}
 \item \textbf{First row.} \textit{``The dynamics of the mirror and its reflections involve the use of a cardboard box. The box is placed on top of a table, and the mirror is positioned in such a way that it reflects the surrounding environment.''}
 \item \textbf{Second row.} \textit{``The mirror's reflection in the cord creates an interesting visual effect, making it appear as if the cord is coming out of the mirror itself. This setup can be useful for individuals who need to charge their electronic devices while working.''}
 \item \textbf{Third row.} \textit{`A pink portable charger is placed on a wooden table next to a mirror. The mirror reflects the portable charger, creating an interesting dynamic between the object and its reflection in the mirror.''}
  \item \textbf{Fourth row.} \textit{``The mirror reflects a pile of dirty clothes or towels, which appear to be wrinkled and disheveled. This phenomenon is caused by the way light bounces off the surface of the mirror and interacts with the objects in front of it''}
\end{itemize}

Text prompts for ``Column-2'' are as follows:
\begin{itemize}
 \item \textbf{First row.} \textit{``Anamorphosis is a technique used to create distorted images that appear normal when viewed from a specific vantage point. In this case, the viewer needs to be positioned directly in front of the mirror to see the full effect of the toy''}
 \item \textbf{Second row.} \textit{``A toy mouse-shaped mirror is placed on a table in front of a yellow cylindrical object. The mirror reflects the environment around it, including the yellow cylinder and other objects in the room.''}
 \item \textbf{Third row.} \textit{``The mirror is reflecting a small button with a cartoon character on it, which is placed on a white surface. The reflection of the button in the mirror creates an interesting visual effect, as the character appears to be floating or hovering''}
  \item \textbf{Fourth row.} \textit{``The mirror in the image is reflecting two toy figures, a red one and a blue one, as they interact with each other. This dynamic creates a playful and imaginative scene, as the toys appear to be having a conversation.''}
\end{itemize}

\paragraph{Figure 8.} Text prompts are as follows:
\begin{itemize}
    \item \textbf{(a)} \textit{``A perfect plane mirror reflection of (A swivel chair with a mesh seat, backrest, and swivel base.) and (A wooden cuboid stool with a square base, slanted squarish seat, and slanted backrest.)''}
    \item \textbf{(b)} \textit{``A perfect plane mirror reflection of (L-shaped sectional sofa with U-shaped backrest, 3D model.) and (A king size bed with a tufted headboard, footboard, slatted base, and a single seater sofa with a backrest and seat cushion.)''}
\end{itemize}

\paragraph{Figure 9.} Text prompts are as follows:
\begin{itemize}
    \item \textbf{(a)} \textit{``A perfect plane mirror reflection of a red and white striped round life buoy surrounded in a cord.''}
    \item \textbf{(b)} \textit{``A perfect plane mirror reflection of a rug''}
\end{itemize}

\section{Additional Results}
\label{sec:addition_res}

\subsection{Affect of Joint training with single and multiple objects}
\label{sec:supp_stage_wise_training_affect}

\begin{table}[!t]
\centering
\caption{Ablation studies on mixed-training for multiple objects}
\label{tab:supp_joint_training}
\begin{adjustbox}{width=\linewidth}
\begin{tabular}{@{}c|ccc|c@{}}
\toprule
%\multicolumn{5}{c}{\textbf{\red{T1} Ablation studies on mixed-training for multiple objects}}             \\ \midrule
Metrics & \multicolumn{3}{c|}{Reflection Generation Quality} & Text Alignment \\ \midrule
Models & \textbf{PSNR} $\uparrow$ & \textbf{SSIM} $\uparrow$ & \textbf{LPIPS} $\downarrow$ & \textbf{CLIP Sim} $\uparrow$ \\ \midrule
\textbf{Joint Training} & 17.41 & 0.615 & 0.153 & \textbf{26.37} \\
\textbf{Ours 50k} & \textbf{18.00} & \textbf{0.744} & \textbf{0.119} & 26.09 \\ \bottomrule
\end{tabular}
\end{adjustbox}

\end{table}

An ablation study in~\cref{tab:supp_joint_training} reveals the significant impact of staged training on generalization. Training on single and multiple splits simultaneously yielded inferior results, highlighting the importance of our staged approach. Curriculum training further allows us to fine-tune on real-world data such as the MSD dataset, providing better results than direct single-stage training as shown in~\cref{sec:ablation_msd_10k} and~\cref{fig:abl_brusnet_zs}.

\subsection{Additional results from single object scenes from~\testsetname{}}
\label{sec:add_res_benchmark_single}

\begin{figure*}[t]
    \centering
    \includegraphics[width=\linewidth]{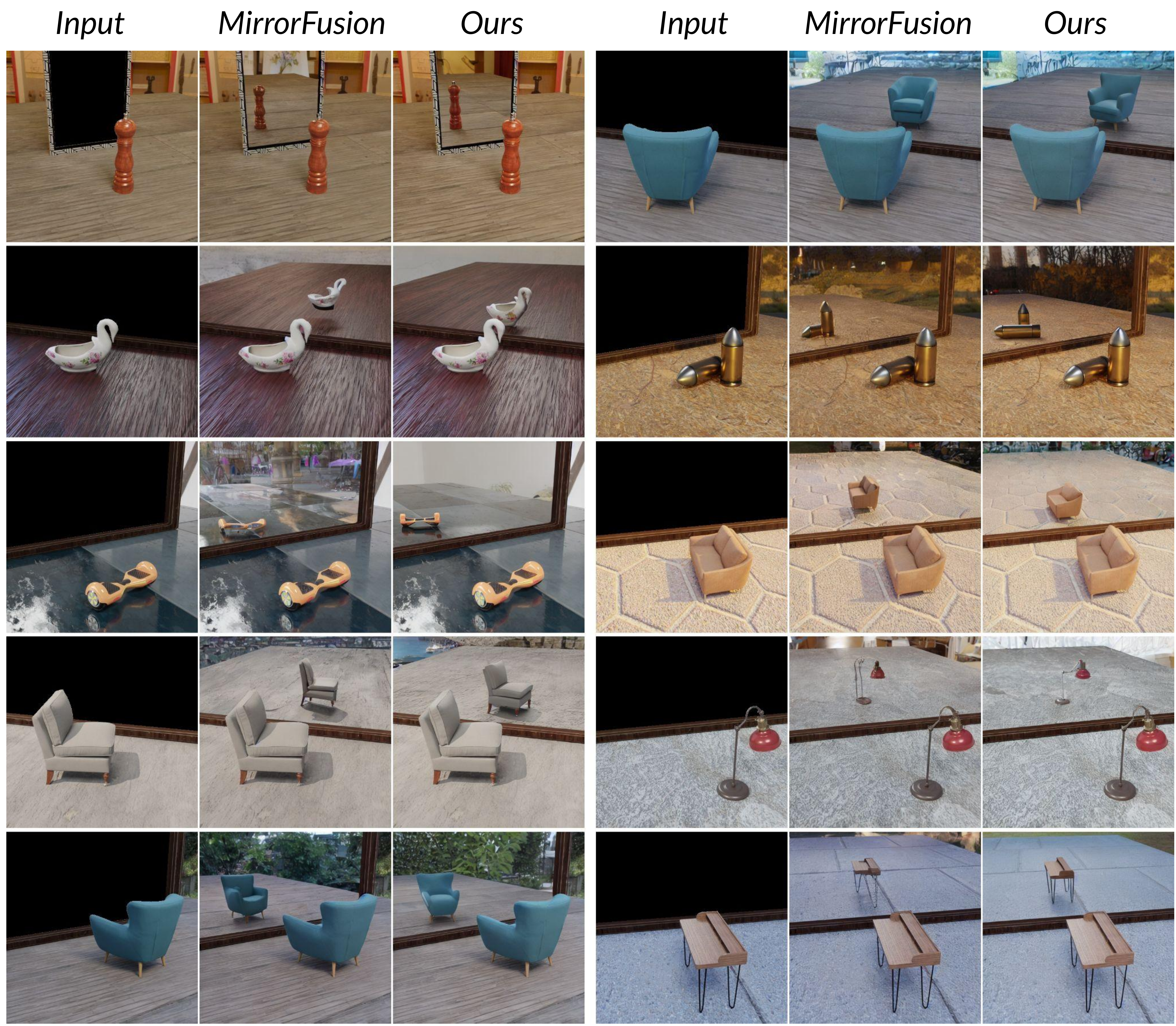}
    \caption{\textbf{Results on scenes with single objects.} More discussion in~\cref{sec:add_res_benchmark_single}.}
    \label{fig:add_res_single_obj}
\end{figure*}

We present additional results for single objects in~\cref{fig:add_res_single_obj}. Observe that the baseline method produces several inaccuracies: the object \textbf{(Column 1, Row 2)} appears to be floating in mid-air with incorrect orientation, and the bullets \textbf{(Column 2, Row 2)} are also misaligned. Additionally, the reflection of the wooden table \textbf{(Column 1, Row 5)} has distorted legs. In contrast, our results accurately capture the geometry and appearance of the object in its reflection.

 Text prompts for \textbf{``Column-1''} in~\cref{fig:add_res_single_obj} are as follows:
\begin{itemize}
 \item \textbf{First row.} \textit{``A perfect plane mirror reflection of a shiny dark wooden pepper mill, stood upright, with a silver ornament on the top.''}
 \item \textbf{Second row.} \textit{``A perfect plane mirror reflection of a white gravy boat with designs of pink flowers on the side and front with green folliage.''}
 \item \textbf{Third row.} \textit{``A perfect plane mirror reflection of a orange and black two wheeled hoverboard.''}
  \item \textbf{Fourth row.} \textit{``A perfect plane mirror reflection of a 3D swivel chair model with a curved backrest, armrests, and a swivel base.''}
  \item \textbf{Fifth row.} \textit{``A perfect plane mirror reflection of a swivel chair with a slender, curved backrest and armrests, featuring a slanted seat.''}
\end{itemize}

Text prompts for \textbf{``Column-2''} in~\cref{fig:add_res_single_obj} are as follows:
\begin{itemize}
 \item \textbf{First row.} \textit{``A perfect plane mirror reflection of a swivel chair with a slender, curved backrest and armrests, featuring a slanted seat.''}
 \item \textbf{Second row.} \textit{``A perfect plane mirror reflection of two metal bullets for either a gun or cannon.''}
 \item \textbf{Third row.} \textit{``A perfect plane mirror reflection of a two-seater sofa with curved backrest, slanted seat, and armrests.''}
  \item \textbf{Fourth row.} \textit{``A perfect plane mirror reflection of a 3D lamp with a cylindrical metal arm, spherical metal base, and spherical glass shade.''}
  \item \textbf{Fifth row.} \textit{``A perfect plane mirror reflection of a rectangular table with a slatted, slanted top, hairpin legs, and a metal frame.''}
\end{itemize}

\subsection{Additional results from multiple object scenes from~\testsetname{}}
\label{sec:add_res_benchmark_multiple}
We present additional results for multiple objects in~\cref{fig:add_res_mul_obj}.
The baseline method struggles to generate accurate reflections in scenes with multiple objects compared to its performance in single-object scenes. Notably, reflections of two sofas \textbf{(Column 1, Row 4)} and a sofa-table \textbf{(Column 2, Row 4)} pair are incorrectly merged, and in some cases, only a single object is rendered in the reflection. This poor performance is primarily due to the limited diversity of the dataset used to train the baseline method. In contrast, our approach preserves the original geometry of the objects, accurately captures their spatial relationships, and maintains their appearance, resulting in significantly more realistic and consistent reflections for scenes with multiple objects.

Text prompts for \textbf{``Column-1''} in~\cref{fig:add_res_mul_obj} are as follows:
\begin{itemize}
 \item \textbf{First row.} \textit{``A perfect plane mirror reflection of a (rug) and (Rectangular cabinet with a slanted roof and base, featuring a door and a drawer, 3D modeled as a TV stand.).''}
 \item \textbf{Second row.} \textit{``A perfect plane mirror reflection of a (3D model of a chair with a backrest, armrests, and seat.) and (Spherical table with a round top, square slanted base, and two slender legs.).''}
 \item \textbf{Third row.} \textit{``A perfect plane mirror reflection of (A 3D model of a chair with a curved, tufted backrest, padded seat, armrests, and a squarish bowl with a matching base and lid.) and (rug).''}
  \item \textbf{Fourth row.} \textit{``A perfect plane mirror reflection of (A 3D model of a two to three-seater sofa with a curved, tufted backrest, armrests, and a footrest.) and (3D model of a three-seater sofa with a curved backrest and armrests.).''}
  \item \textbf{Fifth row.} \textit{``A perfect plane mirror reflection of (Cylindrical stool with a spherical top, square base, slanted seat, and backrest.) and (A cuboid with a base, spherical lid, and slanted top and bottom.).''}
  \item \textbf{Sixth row.} \textit{``A perfect plane mirror reflection of (A cylinder with a spherical base and a spherical shade.) and (3D object: Slatted swivel chair with a curved, slanted X-shaped backrest and seat, featuring armrests and legs.).''}
  \item \textbf{Seventh row.} \textit{``A perfect plane mirror reflection of (3D model of a chaise lounge featuring a curved backrest, cushioned seat, armrests, and footrest, made from a single piece of foam.) and (A 3D object resembling a book with a convex spine, slanted top and bottom edges, and stacked pages.).''}
  \item \textbf{Eigth row.} \textit{``A perfect plane mirror reflection of (Two-seater couch with backrest and armrests, and a tetrahedral box with lid.) and (3D model of a rectangular coffee table with a slanted shelf on top, supported by a slanted frame.).''}
  
\end{itemize}

\begin{figure}[t]
    \centering
    \includegraphics[width=\linewidth]{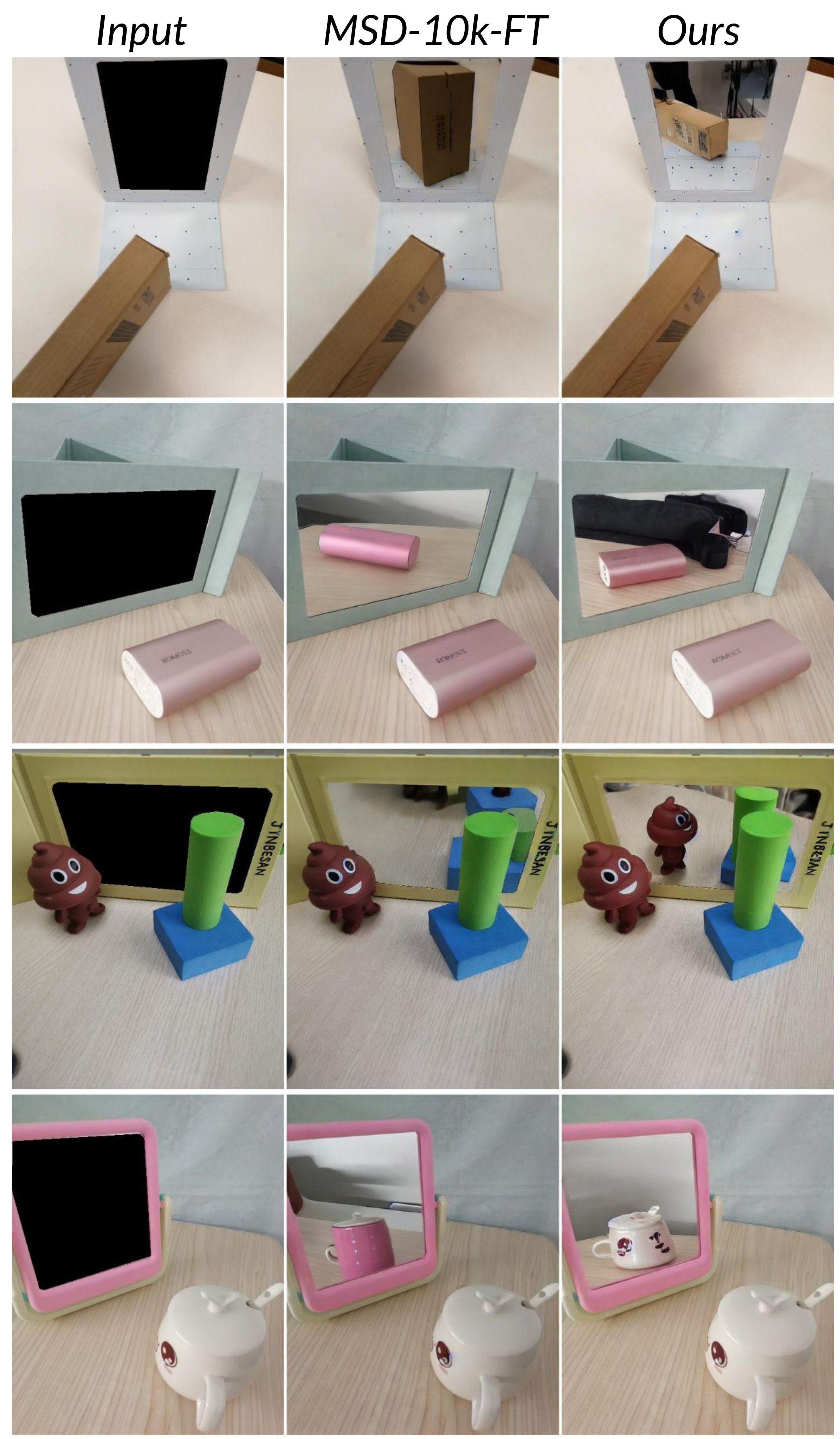}
    \caption{\textbf{Comparison with ``MSD-10k-FT''.} We finetune our model directly on the MSD dataset~\cite{Yang_2019_ICCV} for $10k$ iterations and compare the results with our full $3$-stage finetuning approach. More discussion in~\cref{sec:ablation_msd_10k}.}
    \label{fig:abl_brusnet_zs}
\end{figure}

Text prompts for \textbf{``Column-2''} in~\cref{fig:add_res_mul_obj} are as follows:
\begin{itemize}
 \item \textbf{First row.} \textit{``A perfect plane mirror reflection of (3D model of stacked cylindrical objects with spherical tops, resembling trash cans or water cisterns, featuring a tetrahedral cuboid and a truncated octahedral.) and (Three-drawer dresser with a slanted top, rectangular base, and rectilinear design.).''}
 \item \textbf{Second row.} \textit{``A perfect plane mirror reflection of a (3D model of a chair with a backrest, armrests, and seat.) and (Spherical table with a round top, square slanted base, and two slender legs.).''}
 \item \textbf{Third row.} \textit{``A perfect plane mirror reflection of (rug) and (3D model of a slanted rectangular coffee table in 3ds Max, available for download.).''}
  \item \textbf{Fourth row.} \textit{``A perfect plane mirror reflection of (Three-seater grey sofa with a curved backrest and armrests in 3D.) and (A 3D tetrahedral desk with a slanted top, two drawers, a shelf, and a truncated triangular base.).''}
  \item \textbf{Fifth row.} \textit{``A perfect plane mirror reflection of (A 3D model of a rectangular table with a pair of legs and a top.) and (Two-seater sofa with curved backrest and armrests, 3D model.).''}
  \item \textbf{Sixth row.} \textit{``A perfect plane mirror reflection of (Swivel bar stool with a cylindrical seat, curved backrest, and swivel functionality.) and (Tall cabinet with a triangular base, slanted roof, flat top, and a retractable banner stand.).''}
  \item \textbf{Seventh row.} \textit{``A perfect plane mirror reflection of (A wooden stool with a backrest and a seat.) and (3D model of a lamp with a cylindrical body, spherical base, conical bottom, spherical top, and a cylindrical shade with a spherical accent.).''}
  \item \textbf{Eigth row.} \textit{``A perfect plane mirror reflection of (A king-size platform bed with a box-shaped, curved-top headboard, footboard, side rails, slatted base, and curved backrest.) and (U-shaped sectional sofa with multiple L-shaped sections, featuring backrests and armrests.).''}
  
\end{itemize}

\begin{figure}[t]
    \centering
    \includegraphics[width=\linewidth]{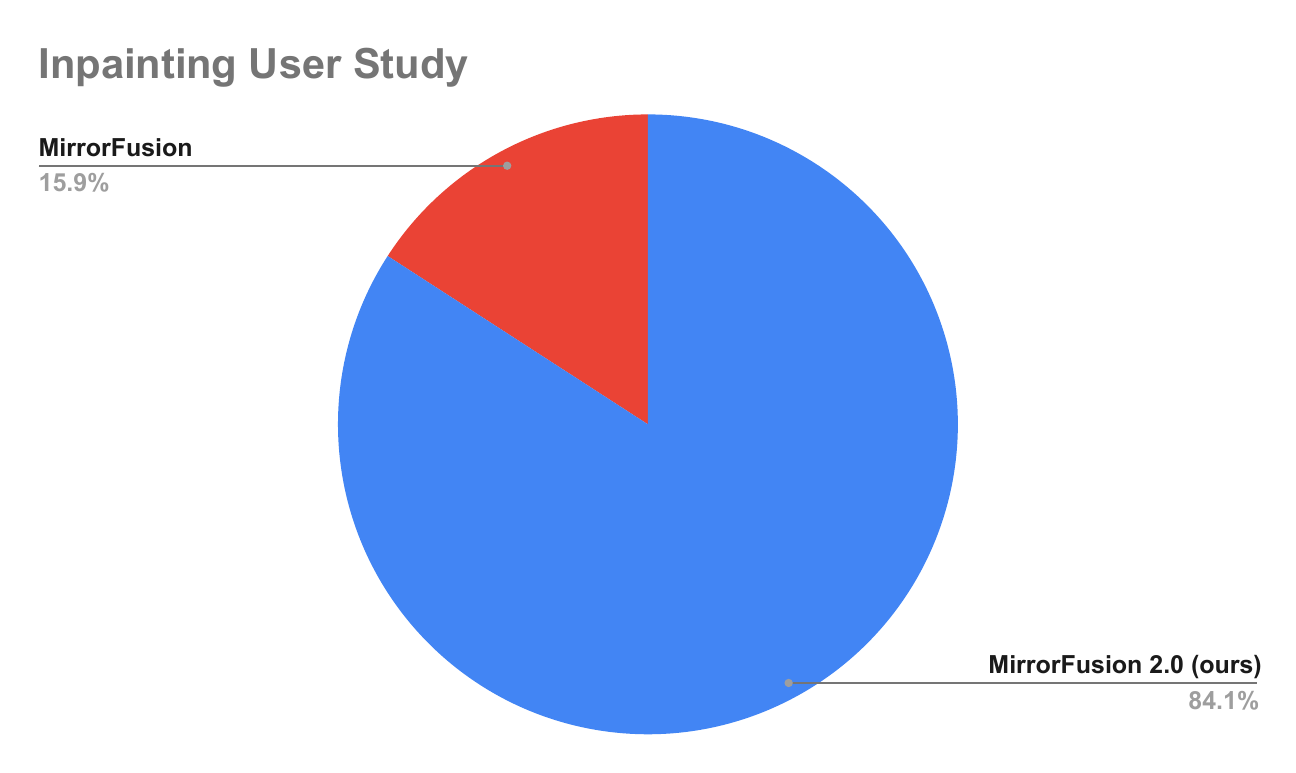}
    \caption{Visual comparison of outputs from our method and the baseline. We discuss in detail in~\cref{sec:supp_user_study}}
    \label{fig:supp_user_study}
\end{figure}

\subsection{Comparison by fine-tuning only on the MSD dataset}
\label{sec:ablation_msd_10k}
To highlight the impact of the proposed dataset~\datasetname{} and stage-wise training more profoundly, we compare our model (which is trained in $3$ stages with the first two involving~\datasetname{}) with the model only finetuned directly on the MSD dataset for $10k$ iterations (i.e only stage 3) from the Stable Diffusion v1.5 checkpoint. We call the fine-tuned model ``MSD-10k-FT''. We compare ``MSD-10k-FT'' with results from our method in~\cref{fig:abl_brusnet_zs}. Note the orientation of the power bank is incorrect \textbf{(Second Row)}, a brown toy is not generated in the reflection \textbf{(Third Row)} and a pink cup is generated instead of the white teapot \textbf{(Fourth Row)} in the results from ``MSD-10k-FT''. This shows the importance of the proposed synthetic dataset for incorporating the priors of accurate mirror reflections and the importance of stage-wise finetuning to bridge the generalization gap.

Text prompts used in Fig.~\ref{fig:abl_brusnet_zs} are as follows:
\begin{itemize}
    \item \textbf{First row.} \textit{``A perfect plane mirror reflection of a cardboard box placed on top of a table''}
    \item \textbf{Second row.} \textit{``A perfect plane mirror reflection of a pink portable charger is placed on a wooden table.''}
    \item \textbf{Third row.} \textit{``A perfect plane mirror reflection of a toy poop emoji figurine placed along with a blue cuboid and a green cylindrical object.''}
    \item \textbf{Fourth row.} \textit{``A perfect plane mirror reflection of a pink and white ceramic mug with a smiling face on it.''}
\end{itemize}

\subsection{User Study Details}
\label{sec:supp_user_study}

We provide details of the user study described in Section 4 of the main paper. We selected 40 samples, including single-object and multi-object scenes, from \testsetname{}, GSO~\cite{downs2022google}, and real-world scenes from MSD~\cite{Yang_2019_ICCV}. These samples were generated by the baseline method~\oldmethod{} and our method~\methodname{}. 

We invited 29 participants (aged 18–50) to compare results based on the realism and plausibility of mirror reflections. Each task involved evaluating and selecting the best result among the outputs from both models with instructions to assess factors such as:
\begin{itemize}
    \item Apparent distance and alignment of objects in the reflection.
    \item Geometry consistency and subtle details in reflections.
    \item Floor reflections and shadow orientations, if present.
\end{itemize}

~\cref{fig:supp_user_study} shows that our method was preferred in 84\% of cases.

\subsection{Comparison with Commercial Product}
\label{sec:app_compare_commercial}
\begin{figure}[t]
    \centering
    \includegraphics[width=\linewidth]{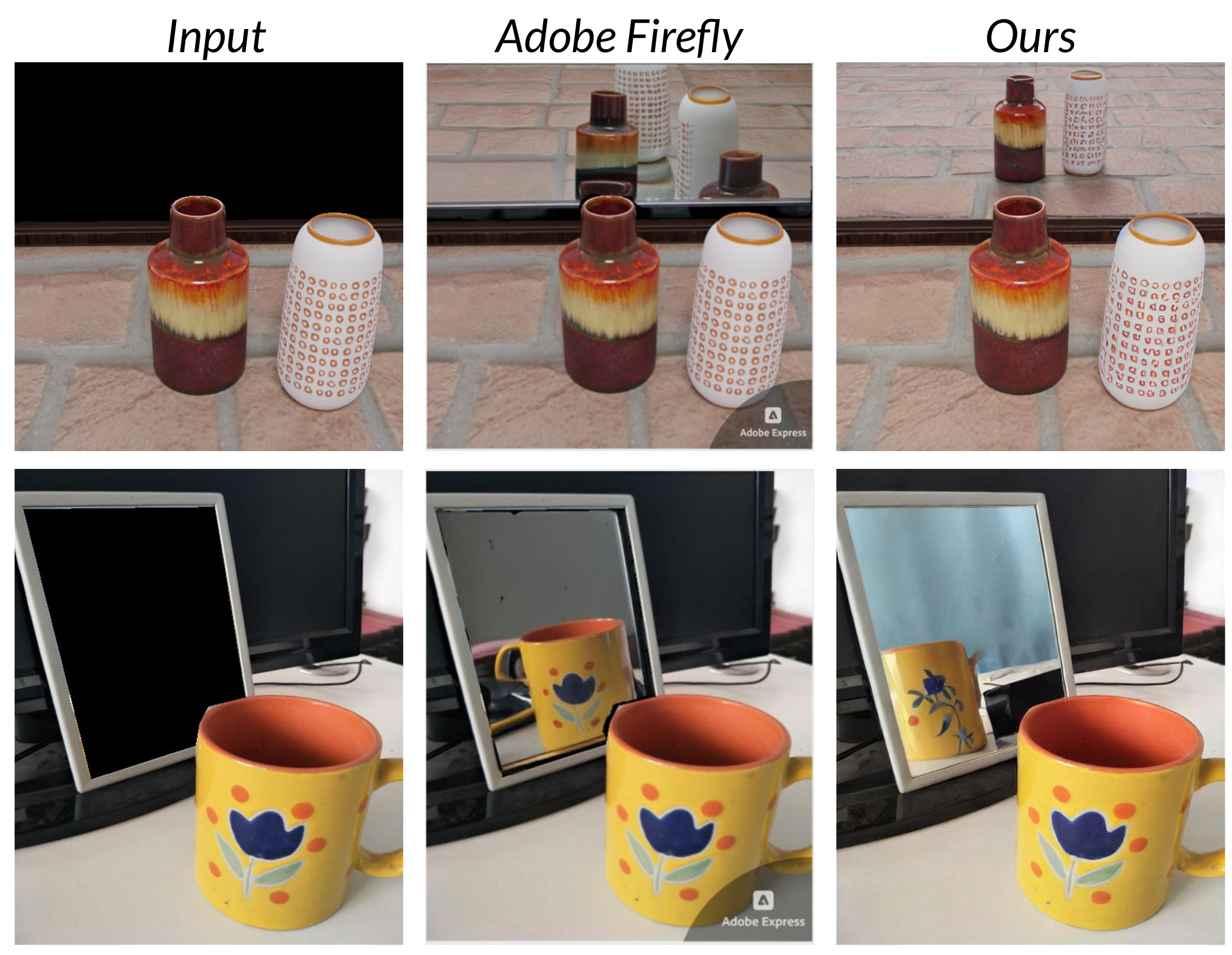}
    \caption{\textbf{Comparison with Adobe Firefly~\cite{adobefirefly}}. We discuss in detail in~\cref{sec:app_compare_commercial}}
    \label{fig:compare_firefly}
\end{figure}
We compare our method with Adobe Firefly~\cite{adobefirefly} in~\cref{fig:compare_firefly}. Our method demonstrates superior performance compared to a widely used commercial product, Adobe Firefly, in generating accurate reflections on a mirror plane. Notably, the commercial product places reflections incorrectly for multiple objects in the scene (top row) and mispositions the reflection of the yellow mug (bottom row). Additionally, the orientation of objects in the inpainted region is inaccurate. In contrast, our method consistently generates reflections in the correct positions while preserving their proper orientation and appearance, resulting in more realistic and visually accurate outcomes. Text prompts used in Fig.~\ref{fig:compare_firefly} are as follows:
\begin{itemize}
    \item \textbf{First row.} \textit{``A perfect plane mirror reflection of (A cylindrical bottle with a spherical top and bottom, featuring a neck and spout.) and (A cylinder with a conical bottom and a spherical top.)''}
    \item \textbf{Second row.} \textit{``A perfect plane mirror reflection of a yellow mug with a flower design is placed on a desk in front of a mirror. The reflection of the mug can be seen in the mirror, creating an interesting visual effect where the mug appears to be floating.''}
\end{itemize}

\begin{figure*}[t]
    \centering
    \includegraphics[width=\linewidth,height=22cm, keepaspectratio]{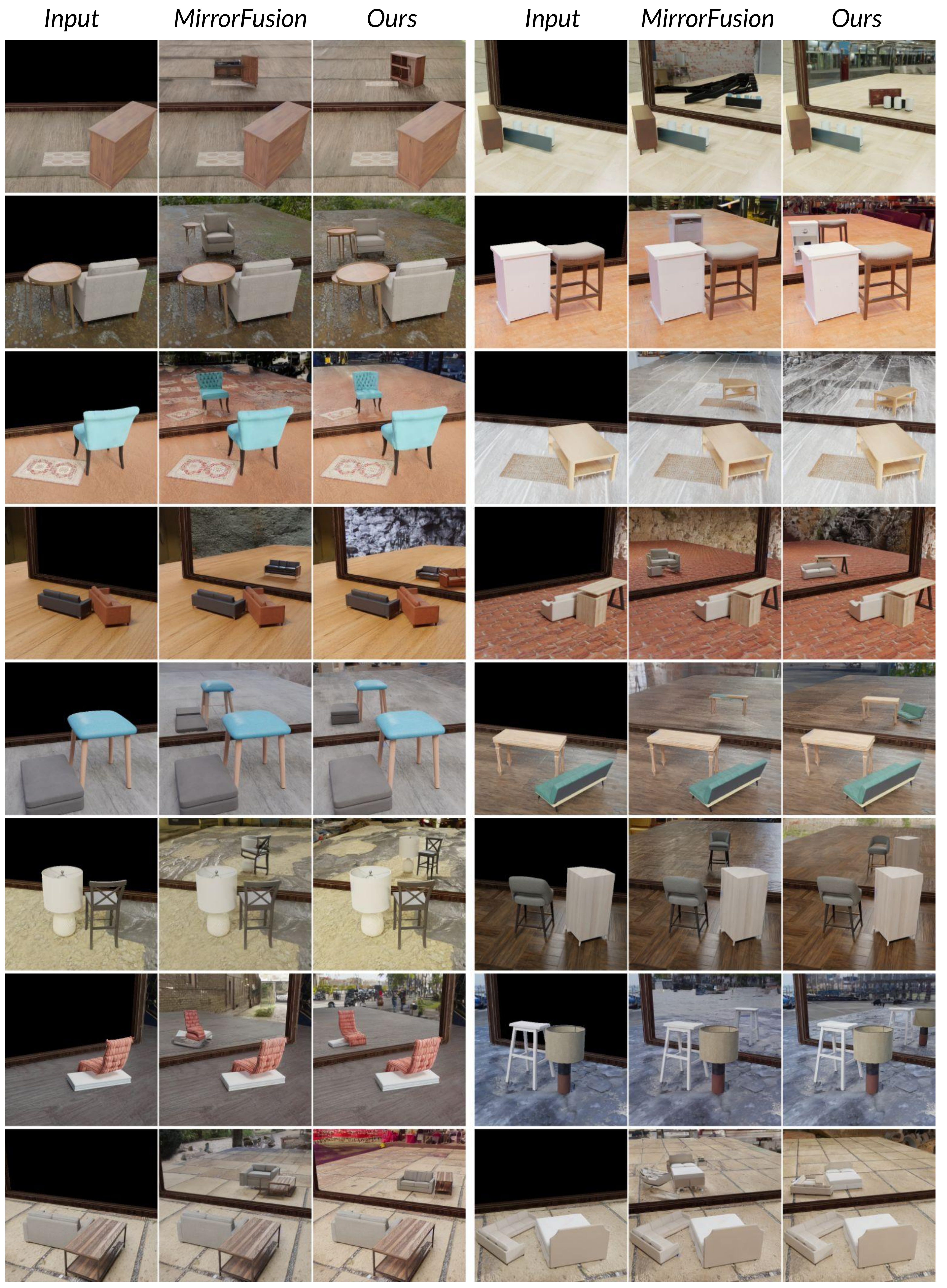}
    \caption{\textbf{Results on scenes with single objects.} More discussion in~\cref{sec:add_res_benchmark_multiple}.}
    \label{fig:add_res_mul_obj}
\end{figure*}